\documentclass[letterpaper,journal]{IEEEtran}

\usepackage{amsmath,amssymb,amsfonts,amsthm}
\usepackage{bm}
\usepackage{cite}
\usepackage[caption=false]{subfig}
\usepackage{textcomp}
\usepackage{stfloats}
\usepackage{url}
\usepackage{verbatim}
\usepackage{cite}
\usepackage{lipsum}
\usepackage{graphicx}
\usepackage{tabularx}
\usepackage{color}

\newtheorem{theorem}{Theorem}

\usepackage{makecell}
\usepackage{array}
\usepackage{longtable}
\usepackage{algorithm}
\usepackage{algorithmic}
\usepackage{color}
\usepackage{multirow}
\usepackage{multicol}
\usepackage{colortbl}
\usepackage{booktabs}
\usepackage{fancyhdr}

\usepackage{pdfpages}

\begin{document}

\title{Zeroth-Order Actor-Critic: An Evolutionary Framework for Sequential Decision Problems}

\author{
Yuheng Lei,
Yao Lyu,
Guojian Zhan,
Tao Zhang,
Jiangtao Li,
Jianyu Chen, \\
Shengbo Eben Li,~\IEEEmembership{Senior Member,~IEEE,}
and Sifa Zheng,~\IEEEmembership{Member,~IEEE}
% <-this % stops a space
\thanks{This study is supported by National Key R\&D Program of China under 2020YFB1600200 and Tsinghua-Toyota Joint Research Fund. \textit{(Corresponding authors: Shengbo Eben Li and Sifa Zheng)}}% <-this % stops a space
\thanks{Yuheng Lei is with the Department of Computer Science, The University of Hong Kong, Hong Kong, China. The majority of this work was done while Yuheng was a master student at School of Vehicle and Mobility, Tsinghua University. (email: leiyh@connect.hku.hk).}% <-this % stops a space
\thanks{Yao Lyu, Guojian Zhan, Shengbo Eben Li and Sifa Zheng are with the School of Vehicle and Mobility, Tsinghua University, Beijing, China (email: lyo.tobias@foxmail.com; zgj21@mails.tsinghua.edu.cn; lishbo@tsinghua.edu.cn; zsf@tsinghua.edu.cn).}% <-this % stops a space
\thanks{Tao Zhang and Jiangtao Li are with SunRising AI Ltd., Beijing, China as chief scientists. (email: zhang.t1983@gmail.com; andy.ljt1988@gmail.com)}
\thanks{Jianyu Chen is with the Institute for Interdisciplinary Information Sciences, Tsinghua University, Beijing, China, and also with the Shanghai Qi Zhi Institute, Shanghai, China (email: jianyuchen@tsinghua.edu.cn).}% <-this % stops a space
\thanks{Code is available at https://github.com/HarryLui98/zoac-tevc.}% <-this % stops a space
}
% The paper headers
\markboth{IEEE TRANSACTIONS ON EVOLUTIONARY COMPUTATION}{}

% \IEEEpubid{0000--0000/00\$00.00~\copyright~2021 IEEE}
% Remember, if you use this you must call \IEEEpubidadjcol in the second
% column for its text to clear the IEEEpubid mark.

\maketitle

%% arxiv version
\thispagestyle{fancy}
\fancyhead{}
\lfoot{\footnotesize{\textcopyright 2025 IEEE. Personal use of this material is permitted. Permission from IEEE must be obtained for all other uses, in any current or future media, including reprinting/republishing this material for advertising or promotional purposes, creating new collective works, for resale or redistribution to servers or lists, or reuse of any copyrighted component of this work in other works.}}
\cfoot{}
\rfoot{}

\begin{abstract}
Evolutionary algorithms (EAs) have shown promise in solving sequential decision problems (SDPs) by simplifying them to static optimization problems and searching for the optimal policy parameters in a zeroth-order way. 
While these methods are highly versatile, they often suffer from high sample complexity due to their ignorance of the underlying temporal structures.
In contrast, reinforcement learning (RL) methods typically formulate SDPs as Markov Decision Process (MDP).
Although more sample efficient than EAs, RL methods are restricted to differentiable policies and prone to getting stuck in local optima.
To address these issues, we propose a novel evolutionary framework Zeroth-Order Actor-Critic (ZOAC).
We propose to use step-wise exploration in parameter space and theoretically derive the zeroth-order policy gradient.
We further utilize the actor-critic architecture to effectively leverage the Markov property of SDPs and reduce the variance of gradient estimators.
In each iteration, ZOAC employs samplers to collect trajectories with parameter space exploration, and alternates between first-order policy evaluation (PEV) and zeroth-order policy improvement (PIM).
To evaluate the effectiveness of ZOAC, we apply it to a challenging multi-lane driving task, optimizing the parameters in a rule-based, non-differentiable driving policy that consists of three sub-modules: behavior selection, path planning, and trajectory tracking. 
We also compare it with gradient-based RL methods on three Gymnasium tasks, optimizing neural network policies with thousands of parameters.
Experimental results demonstrate the strong capability of ZOAC in solving SDPs. ZOAC significantly outperforms EAs that treat the problem as static optimization and matches the performance of gradient-based RL methods even without first-order information, in terms of total average return across all tasks.
\end{abstract}


\vspace{0.15in}

\section{Introduction}

\begin{figure*}[!t]
\centering
\vspace{-0.1in}
\subfloat[EA]{\includegraphics[width=0.24\textwidth, keepaspectratio=true,trim=10 10 30 20,clip]{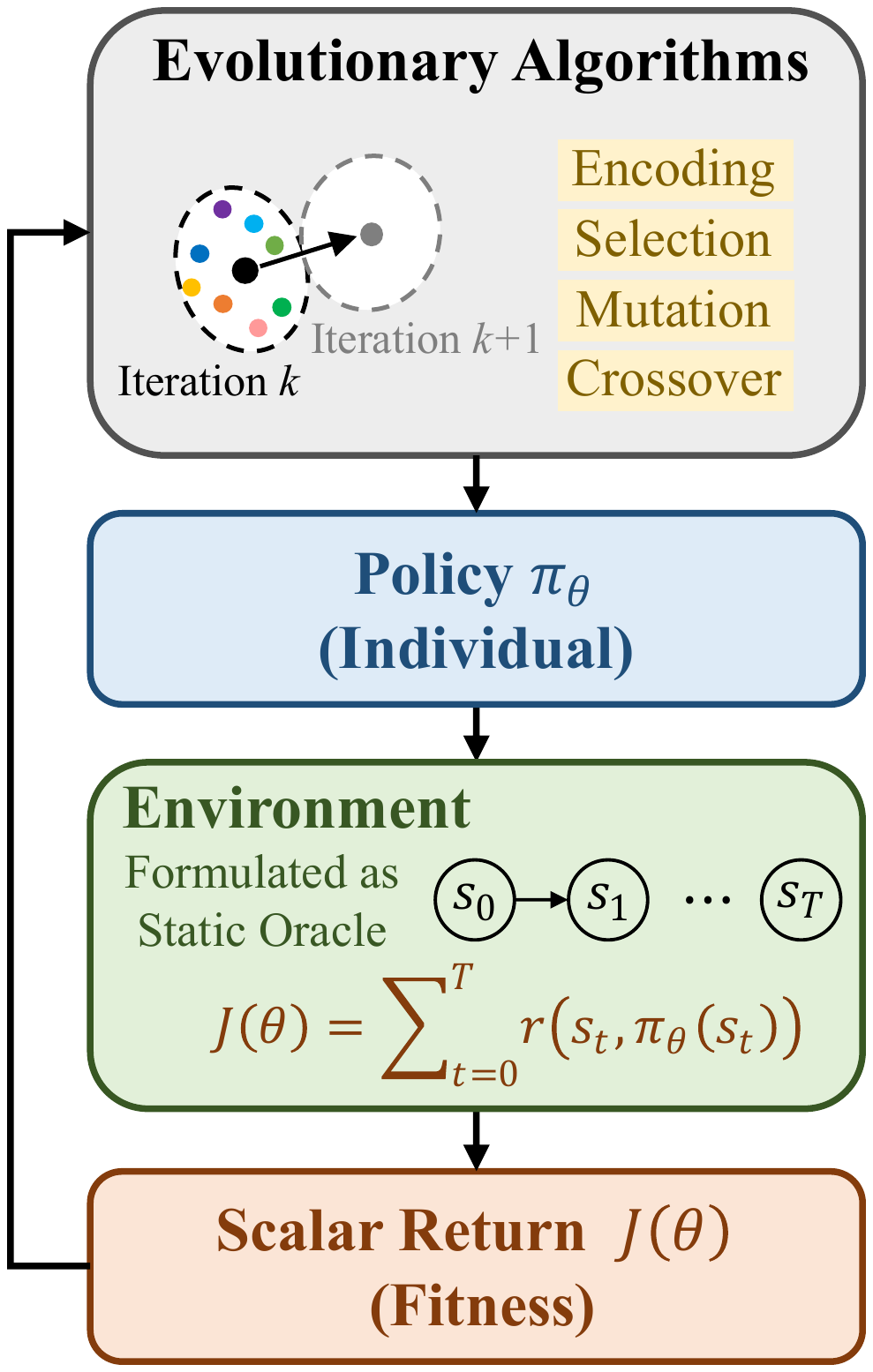}%
\label{fig:intro-ea}}
\hspace{-0.12in}
\subfloat[RL]{\includegraphics[width=0.24\textwidth, keepaspectratio=true,trim=10 10 30 20,clip]{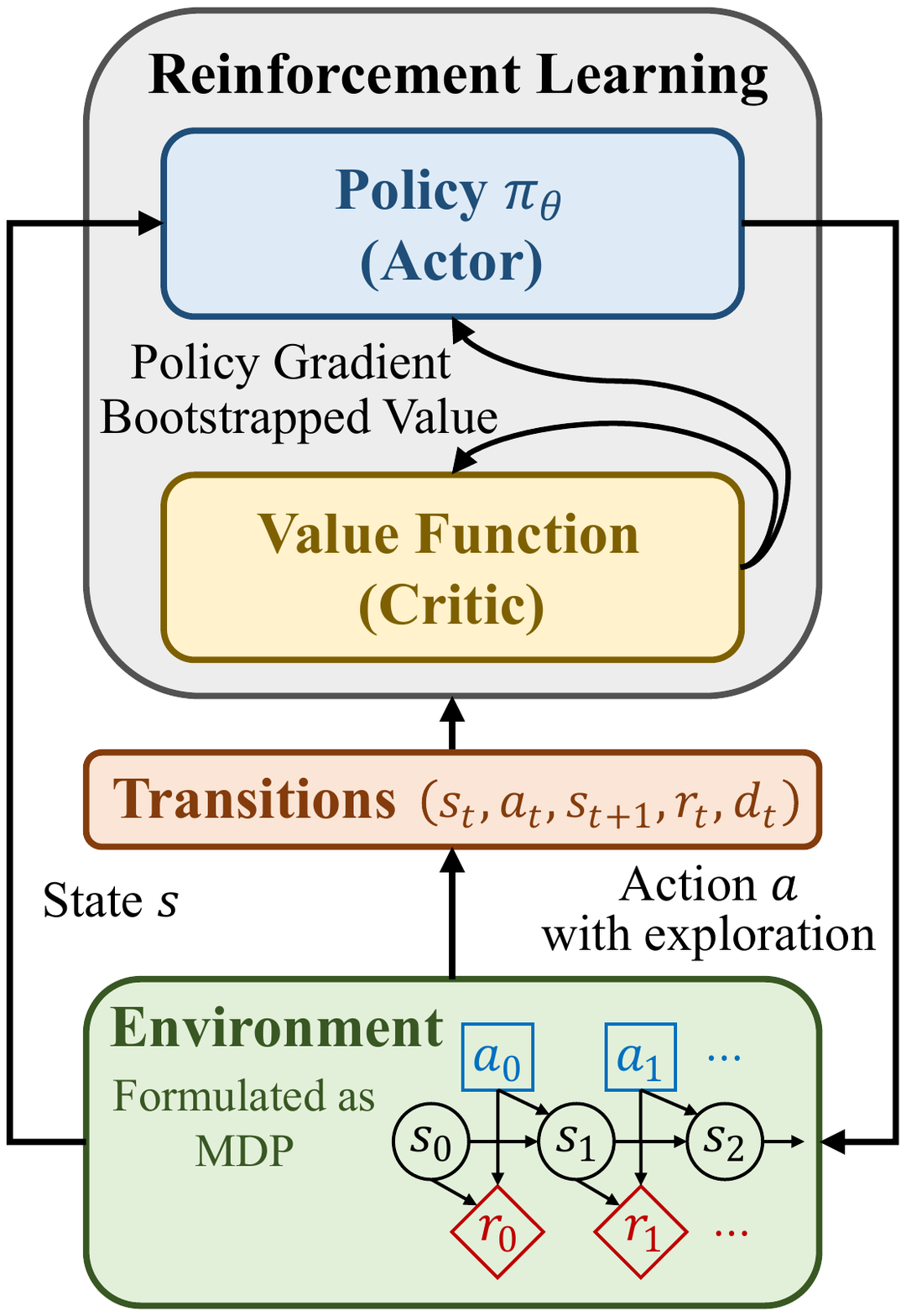}%
\label{fig:intro-rl}}
\hspace{-0.12in}
\subfloat[ZOAC (Proposed)]{\includegraphics[width=0.54\textwidth, keepaspectratio=true,trim=10 16 10 0,clip]{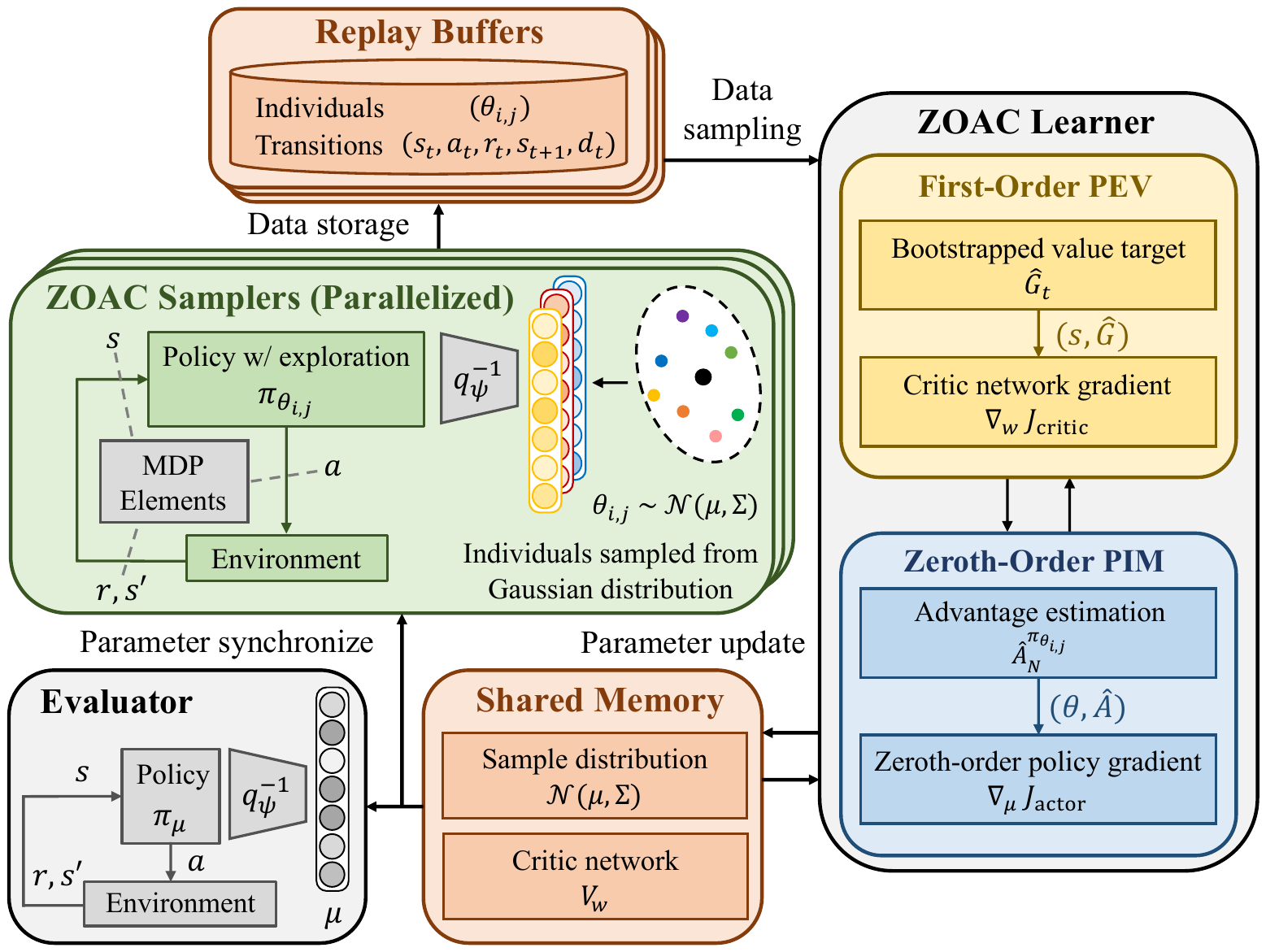}%
\label{fig:intro-zoac}}
\hspace{-0.05in}
\caption{Existing approaches for sequential decision problems: (a) Evolutionary algorithms (EA), (b) Reinforcement Learning (RL). A crucial difference is EA formulates a static optimization problem while RL formulates an MDP with temporal credit assignment. (c) The proposed framework ZOAC, in which samplers collect rollouts with step-wise parameter exploration and the learner alternates between first-order policy evaluation and zeroth-order policy improvement.}
\vspace{-0.15in}
\label{fig:overall}
\end{figure*}

Sequential decision problems (SDPs) provide a general framework to model scenarios where agents need to make a series of decisions in dynamic environments \cite{li2023reinforcement}, such as electronic games \cite{mnih2015human}, autonomous driving \cite{guan2022integrated}, and robotic control \cite{jain2021pixels}. The primary objective in SDPs is to find an optimal decision strategy (i.e., policy) that maximizes the performance metrics over a sequence of time steps. In practice, policies are usually approximated by parameterized functions (e.g., neural networks, fuzzy rules, linear feedback controllers). Consequently, the objective becomes finding the optimal parameter values within the policy parameter space \cite{sigaud2019policy}. 

A straightforward approach is to ignore the underlying temporal structures in SDPs and treat the original dynamic optimization problems as static optimization problems (SOPs). The objective function is the total average return of SDPs, a black-box that can only be queried with noise by sampling trajectories and computing episodic returns \cite{sigaud2019policy, fu2014dynamic}. Therefore, zeroth-order optimization methods (i.e., without using the first-order gradient information) are widely used to solve the SOPs and get the optimal policy of the original SDPs, such as evolutionary algorithms (EAs). The optimization process is illustrated in Figure \ref{fig:intro-ea} and Section \ref{sec:prel.eas}. In the control community, EAs have been used for automatic controller tuning, such as tuning control gains of proportional-integral-derivative (PID) controllers \cite{azar2019implementation, rahayu2022particle, azar2020implementation}, learning the optimal feedback gain matrices \cite{ren2021lqr,malik2019derivative, mohammadi2020linear, mania2018simple, he2023noisy, huang2021accelerated}, optimizing the parameterized model and cost function in model predictive controllers (MPC) \cite{maass2021zeroth, jain2021optimal}. They have also been used to calibrate parameters in expert systems designed for sequential decision problems towards better decision performance, such as fuzzy rules \cite{amer2018adaptive, hein2017particle}, finite state machines \cite{zhang2017finite}, and decision trees \cite{likmeta2020combining, dhebar2022toward}. Recent research has also demonstrated the applicability of EAs in optimizing deep neural networks with over millions of parameters and complex heterogeneous architectures \cite{risi2019deep, jain2021pixels, martinez2022adaptive, such2017deep, song2021esenas, salimans2017evolution}, often referred to as \textit{neuroevolution} \cite{stanley2019designing}.
Despite the wide versatility of EAs, the main limitation lies in sample inefficiency and high variance. Some improved techniques have been proposed from different perspectives, e.g., using orthogonal or antithetic sampling methods \cite{sehnke2010parameter, salimans2017evolution, choromanski2018structured, mania2018simple}, subtracting a baseline for variance reduction \cite{sehnke2010parameter, grathwohl2018backpropagation}, or identifying a low-dimensional subspace \cite{maheswaranathan2019guided, choromanski2019complexity, sener2020learning}. Nevertheless, these methods still treat policy search as static optimization problems and therefore suffers from high sample complexity caused by disregarding the temporal nature of sequential decision making.

From the other side, reinforcement learning (RL) \cite{li2023reinforcement,sutton2018reinforcement} formulate sequential decision problems as Markov decision process (MDP) and has achieved great success in a wide range of challenging tasks. Actor-critic methods, e.g., A3C \cite{mnih2016asynchronous} and PPO \cite{schulman2017proximal}, are among the most popular RL algorithms. As shown in Figure \ref{fig:intro-rl} and detailed in Section \ref{sec:prel.rl}, actor-critic methods consider the temporal nature of sequential decision problems and estimate the state values of collected samples via bootstrapping. These value estimates are then used to construct advantage functions, resulting in much lower variance than the Monte Carlo episodic return used in vanilla policy gradient \cite{williams1992simple}, leading to accelerated learning speed and improved stability. However, existing gradient-based RL algorithms need to differentiate through the entire policy, which may not be applicable when non-differentiable policy structure is involved. In such cases, zeroth-order optimization methods like EAs may be the only inefficient option available. In fact, some recent research have tried to combine EAs and actor-critic methods into hybrid methods, e.g., run EAs in parallel with off-policy actor-critic algorithms and optimize the population of policies with information from both sides \cite{khadka2018evolution, pourchot2018cem, bodnar2020proximal}, or inject parameter noise into existing RL algorithms for efficient exploration \cite{plappert2018parameter, fortunato2018noisy}. However, existing hybrid methods still rely on first-order gradient-based policy improvement, which necessitates a differentiable policy and consequently lacks the generalization capabilities of pure evolutionary algorithms. Our motivation is to improve evolutionary algorithms for solving sequential decision problems by incorporating ideas from actor-critic methods that enhance sample efficiency and training stability, while preserving the zeroth-order policy update mechanism for high parallelization and wide versatility.

In this paper, we propose a novel framework Zeroth-Order Actor-Critic algorithm (ZOAC) that unifies temporal difference learning and evolutionary computation into an actor-critic architecture to take advantage of their complementary strengths. We first propose to replace the episode-wise parameter space exploration used in EAs with a step-wise one to encourage sufficient exploration. We then derive the unbiased zeroth-order policy gradient with respect to the Gaussian smoothed objective under this scheme. To leverage the temporal structure of sequential decision problems, we further integrate a critic network to approximate the state-value function and construct advantage estimators for policy gradients. The practical implementation of ZOAC utilizes the open-source distributed computing framework Ray \cite{moritz2018ray} and consists of samplers, evaluators, replay buffers, and a learner. Figure \ref{fig:intro-zoac} provides an overview of the system architecture. In each iteration, parallelized samplers collect rollouts and store transitions in replay buffers. The learner alternates between first-order policy evaluation (PEV) that update the critic network and zeroth-order policy improvement (PIM) that update the actor. The evaluator periodically evaluates the policy and record evaluation metrics. To validate the effectiveness of ZOAC, we apply it to a challenging multi-lane autonomous driving task. We design a rule-based autonomous driving decision and control policy with three sub-modules: behavior selection, path planning, and trajectory tracking, containing 27 trainable parameters in total. We also compare it with gradient-based RL methods on the standard Gymnasium benchmark \cite{towers2024gymnasium}, optimizing neural network policies with thousands of parameters. Experimental results demonstrate the excellent performance of ZOAC in terms of sample efficiency and training stability. Notably, ZOAC outperforms evolutionary algorithms that handle the problem as static optimization and achieves performance on par with gradient-based RL methods, even without first-order information, in terms of total average return across all tasks. The main contribution of this study are summarized as follows:
\begin{enumerate}
\item A novel zeroth-order actor-critic evolutionary framework is proposed to tackle sequential decision problems. This framework inherits the derivative-free nature and can be applied to arbitrary parameterized policies, while achieving higher sample efficiency and better training stability by leveraging the temporal structure.
\item A policy training system utilizing parallel computing architecture and the proposed ZOAC framework has been established and extensively tested on a rule-based autonomous driving policy training problem. Experimental results demonstrate the effectiveness of our proposed framework over existing baselines that simply formulate the problem as static optimization.
\end{enumerate}

The remainder of this article is organized as follows. Section \ref{sec:prel} overview the fundamentals on reinforcement learning and evolutionary algorithms. Section \ref{sec:framework} describes the proposed ZOAC framework, from theoretic analysis to practical implementation. Section \ref{sec:problem} illustrates the autonomous driving policy training problem, while Section \ref{sec:exp} presents and analyzes the experimental results. Section \ref{sec:conclusion} gives the conclusion of the study and discusses future research.

\section{Preliminaries and Notations}
\label{sec:prel}

\subsection{Sequential Decision Problems as Markov Decision Process}
\label{sec:prel.sdp}

Sequential Deicision problems can be formulated as a \textit{Markov Decision Process} (MDP) defined as $(\mathcal{S}, \mathcal{A}, \mathcal{P}, r)$, where $\mathcal{S}$ is the state space, $\mathcal{A}$ is the action space,
%and they contain all possible states $s$ and actions $a$, respectively.
$\mathcal{P}: \mathcal{S}\times\mathcal{A}\times\mathcal{S}\to\mathbb{R}$ is the transition probability matrix, $r:\mathcal{S}\times\mathcal{A}\to\mathbb{R}$ is the reward function. In an MDP, the conditional probability of the next state depends only on the present state and action, i.e., the state transitions of an MDP satisfy the \textit{Markov property}.

The agent's behavior is defined as a policy $\pi:\mathcal{S}\to\mathcal{A}$, which maps from the state space to the action space. At each time step $t$, the agent receives the current state $s_t$ and selects an action according to the policy $a_t=\pi(s_t)$. The system transitions to the next state $s_{t+1}$ with a probability of $p(s_{t+1}|s_{t},a_{t})$ according to the system dynamics, and receives a reward signal $r_t=r(s_{t}, a_{t})$. In an unambiguous situation, $s$ and $s'$ is also used to refer to the current state and the next state respectively.
The optimization objective in an MDP is to find an optimal policy $\pi^*$ that either maximizes (1) average reward objective $J_\pi^{\text{Avg}}$, the expectation of the reward per stage under the stationary state distribution $d_\pi$ or (2) discounted reward objective $J_\pi^{\text{Dis}}$ with $\gamma\in[0,1)$ as the discount factor, the expectation of state-value function under the initial state distribution $d_0$. The objective can be formally represented as:
\begin{align}
\label{eq:prel-objpi-avg}
J_\pi^{\text{Avg}}&\!=\!\lim_{N\to\infty}\!\frac{1}{N}\mathbb{E}_{\pi}\Big[\!\sum_{t=0}^{N-1} r(s_t,a_t)\Big]\!\!=\!\mathbb{E}_{s\sim d_\pi,a\sim\pi}\left[r(s,a)\right],\\ \label{eq:prel-objpi-dis}
J_\pi^{\text{Dis}}&\!=\!\mathbb{E}_{s_0\sim d_0,\pi}\Big[\!\sum_{t=0}^{\infty}\gamma^t r(s_t,a_t)\Big]\!\!=\!\mathbb{E}_{s\sim d_0}[v^{\pi}(s)],
\end{align}

\vspace{-0.1in}
\subsection{Reinforcement Learning and Actor-Critic Methods}
\label{sec:prel.rl}

Reinforcement Learning (RL) \cite{sutton2018reinforcement, li2023reinforcement} is a widely used approach for solving MDP \eqref{eq:prel-objpi-avg} or \eqref{eq:prel-objpi-dis}. The policy iteration framework usually consists of two alternating steps: \textit{policy evaluation} (PEV) and \textit{policy improvement} (PIM). In practice, parameterized models are often used to approximate the optimal policy (actor) and value function (critic). The corresponding actor-critic architecture is shown in Figure \ref{fig:intro-rl}.

PEV aims to estimate the value function for a given policy using, e.g., Monte Carlo method (MC) or temporal difference methods (TD). Compared to MC, TD fully utilizes the Markov property of the sequential decision problem and significantly reduces the variance of state value estimation. For example, in discounted reward settings, it learns the state-value function in a bootstrapping way by reusing the existing value estimates based on the self-consistency equations \cite{li2023reinforcement}:
\begin{align}
\label{eq:prel-selfconsisv}
v^\pi(s)&=\mathbb{E}_{s'\sim\mathcal{P}}[r(s,\pi(s))+\gamma v^\pi(s')],\\ \label{eq:prel-selfconsisq}
q^\pi(s,a)&=\mathbb{E}_{s'\sim\mathcal{P}}[r(s,a)+\gamma q^\pi(s',\pi(s'))].
\end{align}
Besides, advantage function $A^\pi(s,a)=q^\pi(s,a)-v^\pi(s)$ is often introduced to describes how much better it is to take a specific action $a$ in state $s$ over using policy $\pi$.

PIM aims to improve the policy based on the estimated values. For any differentiable parameterized policy $\pi_{\boldsymbol{\theta}}$, the deterministic policy gradient theorem holds \cite{silver2014deterministic}:
\begin{equation}
\label{eq:dpg}
\nabla_{\boldsymbol{\theta}} J(\boldsymbol{\theta})=\mathbb{E}_{s\sim d_{\pi_\theta}}[\nabla_{\boldsymbol{\theta}}\pi_{\boldsymbol{\theta}}(s)\nabla_a Q^{\pi_{\boldsymbol{\theta}}}(s, a)\vert_{a=\pi_{\boldsymbol{\theta}}(s)}]
\end{equation}
where $d_{\pi_\theta}$ is the discounted state distribution. If we denote the probability of transitioning from state $s$ to state $s'$ after $t$ time steps under policy $\pi_{\boldsymbol{\theta}}$ as $p(s\rightarrow s', t, \pi_{\boldsymbol{\theta}})$, then
\begin{equation}
\label{eq:prel-distribution}
d_{\pi_{\boldsymbol{\theta}}}(s')=\int_{\mathcal{S}}\sum_{t=0}^{\infty}\gamma^t d_0(s)p(s\rightarrow s', t, \pi_{\boldsymbol{\theta}})\mathrm{d}s.
\end{equation}

A common choice in deep reinforcement learning (DRL) is to use neural networks (NNs) and optimize them with first-order optimization techniques, e.g., stochastic gradient descent (SGD). However, since such black-box approximators are extremely difficult to interpret and formally analyze, rule-based policies that benefit from rich theories and excellent interpretabilities are still preferred in many scenarios. Due to the prevailing non-differentiability of such policies with respect to their design parameters, existing first-order gradient-based RL methods are usually not applicable in such cases.

\vspace{-0.1in}
\subsection{Evolution Strategies}
\label{sec:prel.eas}

Evolutionary algorithms (EAs) refer to a class of algorithms that simulate natural biological evolution mechanisms. The optimization process in EAs typically involves the following steps until the optimal solution is found: (1) a population of individuals is randomly generated, each of them representing a set of parameters, often called \textit{genotypes}; (2) each individual in the population is evaluated using a \textit{fitness} function, which quantifies how well the individual solves the problem; (3) according to the fitness values, individuals are \textit{selected} and \textit{mutated} to generate a new population for the next iteration.

Evolutionary strategy (ES) is one of the most popular evolutionary algorithms, which uses a probability distribution to describe the distribution of individuals in the population. During optimization, the distribution is updated based on the fitness values of each individual, gradually moving towards the optimal solution. Remarkable examples include Natural ES (NES) \cite{wierstra2014natural}, Cross Entropy Method (CEM) \cite{rubinstein2004cross}, and Covariance Matrix Adaptation ES (CMA-ES) \cite{hansen2001completely}.

Let $f:\mathbb{R}^D\to\mathbb{R}$ be the function we want to optimize, and only zeroth-order information (i.e., the function value itself) is available. ES optimizes the Gaussian smoothed objective $J^{\textnormal{(ES)}}({\boldsymbol{\mu},\boldsymbol{\Sigma}})=\mathbb{E}_{\boldsymbol{\theta}\sim\mathcal{N}(\boldsymbol{\mu},\boldsymbol{\Sigma})}f(\boldsymbol{\theta})$,
where $\boldsymbol{\mu}$ and $\boldsymbol{\Sigma}$ is the mean and covariance matrix of the Gaussian distribution respectively. The probability density function of Gaussian distribution can be represented as:
\begin{equation*}
\label{eq:prel-gaussianpdf}
p(\boldsymbol{\theta}|\boldsymbol{\mu}, \boldsymbol{\Sigma})\!=\!\frac{1}{(2\pi)^{D/2}|\boldsymbol{\Sigma}|^{1/2}} \exp\!\left[-\frac{1}{2}(\boldsymbol{\theta}-\boldsymbol{\mu})^T \boldsymbol{\Sigma}^{-1} (\boldsymbol{\theta}-\boldsymbol{\mu})\right]\!.
\end{equation*}
and the log-likelihood derivatives can be calculated as:
\begin{align}
\boldsymbol{w}_{\boldsymbol{\mu},\boldsymbol{\Sigma}}(\boldsymbol{\theta})&\triangleq\frac{\partial \log p(\boldsymbol{\theta}|\boldsymbol{\mu}, \boldsymbol{\Sigma})}{\partial \boldsymbol{\mu}}=\boldsymbol{\Sigma}^{-1}(\boldsymbol{\theta} - \boldsymbol{\mu}),
\end{align}

The ES-style zeroth-order gradient with respect to the mean $\boldsymbol{\mu}$
% and the covariance matrix $\boldsymbol{\Sigma}$ 
can be derived using the log-likelihood ratio trick:
\begin{equation}
\begin{aligned}
\label{eq:prel-esgrad.mean}
\!\!\!\nabla_{\boldsymbol{\mu}}J^{\textnormal{(ES)}}({\boldsymbol{\mu},\boldsymbol{\Sigma}})&=\nabla_{\boldsymbol{\mu}}\int f(\boldsymbol{\theta})p(\boldsymbol{\theta}|\boldsymbol{\mu}, \boldsymbol{\Sigma})\mathrm{d}\boldsymbol{\theta}\\
&=\int f(\boldsymbol{\theta})p(\boldsymbol{\theta}|\boldsymbol{\mu}, \boldsymbol{\Sigma})\nabla_{\boldsymbol{\mu}}\log p(\boldsymbol{\theta}|\boldsymbol{\mu}, \boldsymbol{\Sigma})\mathrm{d}\boldsymbol{\theta}\\
&=\mathbb{E}_{\boldsymbol{\theta}\sim\mathcal{N}(\boldsymbol{\mu},\boldsymbol{\Sigma})}[f(\boldsymbol{\theta})\boldsymbol{w}_{\boldsymbol{\mu},\boldsymbol{\Sigma}}(\boldsymbol{\theta})].
\end{aligned}
\end{equation}

In practice, the expectation of the Gaussian distribution is estimated through Monte Carlo sampling. We sample $n$ individuals with \textit{genotypes} $\{\boldsymbol{\theta}_i\}_{i=1,2,\cdots,n}$ in each iteration. For each individual, we can calculate the corresponding log-likelihood derivatives $\boldsymbol{w}_{\boldsymbol{\mu},\boldsymbol{\Sigma}}(\boldsymbol{\theta}_{i})$. 
When applying ES in sequential decision problems, existing arts \cite{salimans2017evolution, mania2018simple, jain2021pixels, o2020tunercar, jain2021optimal, hansen2009method, he2023noisy, huang2021accelerated} usually regard the episodic return as the \textit{fitness} function, i.e., $f(\boldsymbol{\theta})=\mathbb{E}_{s\sim d_0}\left[v^{\pi_{\boldsymbol{\theta}}}(s)\right]$. As shown in Figure \ref{fig:intro-ea}, the sequential decision problem actually degenerates into a static optimization problem, in which querying the fitness function for one time is equivalent to sampling a trajectory and computing the episodic return. Then the zeroth-order gradient estimates can be computed as:
\begin{align}
\label{eq:prel-esgradest.ret}
f^{\text{(ES)}}_i&=\hat{G}_{i,0}^{\pi_{\boldsymbol{\theta}_i}}=\sum_{j=0}^{NH}\gamma^{j}r_{i,j},\\ \label{eq:prel-esgradest.mean}
\nabla_{\boldsymbol{\mu}}{\hat{J}^{\text{(ES)}}}&\approx\frac{1}{n}\sum_{i=1}^{n}f^{\text{(ES)}}_{i}\boldsymbol{w}_{\boldsymbol{\mu},\boldsymbol{\Sigma}}(\boldsymbol{\theta}_{i}).
\end{align}

ES offers the advantage of easy parallelization (e.g., enabling almost linear speedup in the number of CPU cores \cite{salimans2017evolution, huang2021accelerated}), which allows for significantly reduced wall-clock time. On the other hand, ES works with full-length episodes to obtain fitness values, resulting in high sample complexity.

\section{Proposed Framework}
\label{sec:framework}
In this section, we aim to improve the sample efficiency and training stability of ES, while inheriting its advantages such as easy parallelization and high versatility. We start from generalizing
the episode-wise parameter mutation used in ES to a step-wise one, which encourages sufficient exploration. From this foundation, we derive the zeroth-order policy gradient that considers the Markov property. We finally propose the Zeroth-Order Actor-Critic framework (ZOAC) shown in Figure \ref{fig:intro-zoac}. By alternating between first-order policy evaluation (PEV) and zeroth-order policy improvement (PIM), it can effectively leverage the temporal nature of sequential decision problems and significantly accelerate the training process.

\vspace{-0.1in}
\subsection{Step-wise Exploration in Parameter Space}
\label{sec:framework.exploration}

\begin{figure}[t]
\centering
\includegraphics[width=0.93\linewidth, keepaspectratio=true,trim=120 100 120 100,clip]{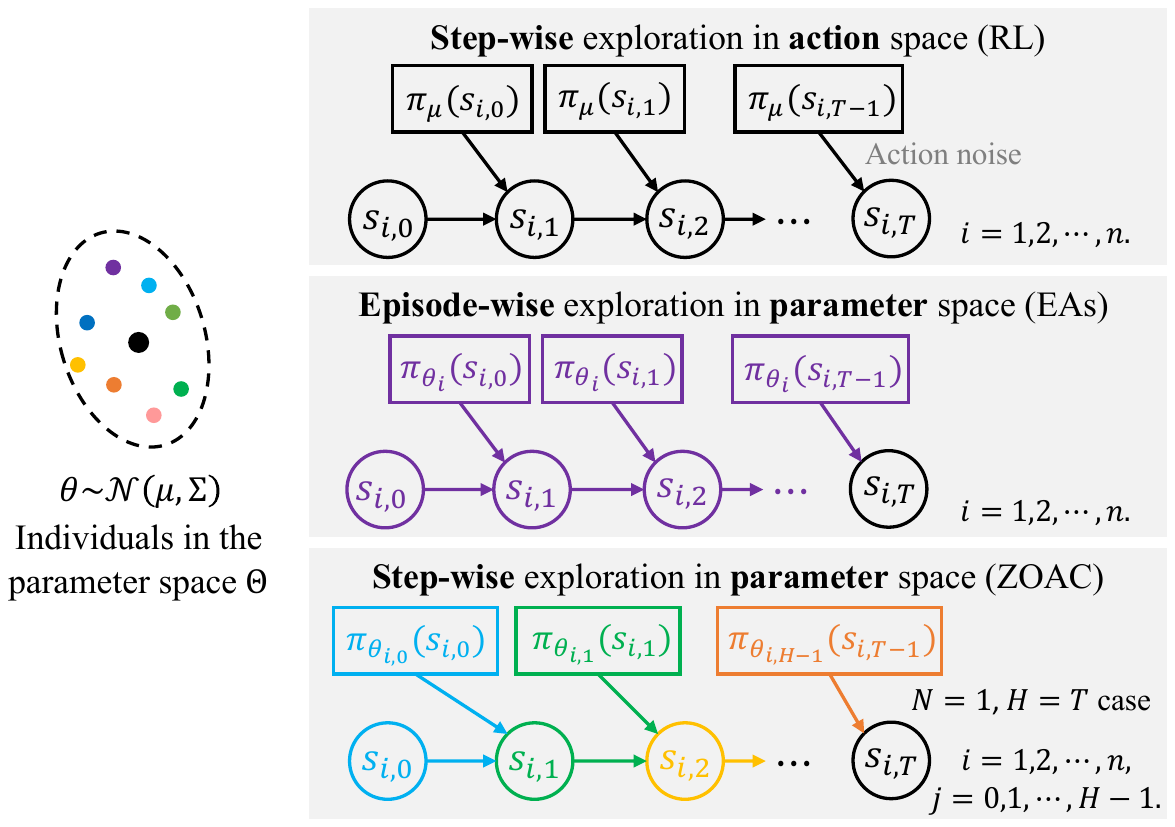}%
\vspace{-0.1in}
\caption{Comparison of exploration strategies during sampling.}
\vspace{-0.2in}
\label{fig:method-exploration}
\end{figure}

Exploitation-exploration trade-off has been a long-standing challenges in reinforcement learning\cite{sutton2018reinforcement, li2023reinforcement}, focusing on how to gather more diverse data and reduce the chance of getting stuck in local optima. Figure \ref{fig:method-exploration} compares common exploration strategies during sampling. Existing gradient-based reinforcement learning methods optimize a single policy but use step-wise action noise to explore, e.g., $\epsilon$-greedy strategy \cite{mnih2015human}, Ornstein-Uhlenbeck process noise \cite{lillicrap2015continuous}, Gaussian noise \cite{fujimoto2018addressing}. On the other hand, evolutionary algorithms maintain a population of policies (with different parameters) for sampling, resulting in more complex and coherent exploration and producing diverse behavioral patterns \cite{fortunato2018noisy, plappert2018parameter, raffin2022smooth}. In the episode-wise parameter space exploration, each perturbed policy is determined before trajectory starts and remains unchanged throughout the entire trajectory. However, this approach can lead to a dilemma. Consider the ES-style gradient estimates \eqref{eq:prel-esgradest.mean}, if a large number of individuals $n$ is evaluated, the sample complexity will increase significantly, especially when solving long-horizon problems. Conversely, when $n$ is small, the zeroth-order gradient estimated as the weighted sum of several random directions exhibits excessively high variance \cite{berahas2021theoretical}, which can greatly harm the overall performance.

To address this issue, we propose a novel step-wise parameter space exploration strategy, which $n$ trajectories are sampled in each iteration and in total $nH$ individuals in the parameter space are evaluated. Each parameter noise sample explores a fixed number of steps, denoted as $N$. To be specific, if we denote the state at the $t$-th step in the $i$-th trajectory sampled in a certain iteration as $s_{i,t}$, the exploration strategy can be described as: when reaching states that can be represented as $s_{i,jN}$ where $j\in\mathbb{N}$, an individual $\boldsymbol{\theta}_{i,j}$ is sampled from the Gaussian distribution $\mathcal{N}(\boldsymbol{\mu},\boldsymbol{\Sigma})$ and the corresponding deterministic policy $\pi_{\boldsymbol{\theta}_{i,j}}$ is used to collect samples for $N$ steps. The bottom part of Figure \ref{fig:method-exploration} illustrate the case when $N=1$, in which the policy parameter is sampled identically and independently at every steps. We can regard the proposed exploration strategy as a stochastic policy $\beta_{\boldsymbol{\mu},\boldsymbol{\Sigma}}$. A limiting case is when $N$ grows to the episode length, the proposed strategy reverts back to the episode-wise one.

We further introduce a notation $\boldsymbol{\xi}={\boldsymbol{\Sigma}}^{-\frac{1}{2}}(\boldsymbol{\theta}-\boldsymbol{\mu})$ for brevity. This transformation leads to $\boldsymbol{\xi}\sim \mathcal{N}(\boldsymbol{0},\boldsymbol{I}_D)$, and we also have $\boldsymbol{\theta}=\boldsymbol{\mu}+{\boldsymbol{\Sigma}}^{\frac{1}{2}}\boldsymbol{\xi}$. Note that ${\boldsymbol{\Sigma}}^{\frac{1}{2}}$ is the (unique) principal square root matrix of the covariance matrix $\boldsymbol{\Sigma}$. In other words, each $\boldsymbol{\xi}$ has a one-to-one correspondence with $\boldsymbol{\theta}$, and we will use them interchangeably in the remainder of this paper.

\vspace{-0.1in}
\subsection{Zeroth-order Policy Gradient}
\label{sec:framework.zopg}
\subsubsection{Optimization Objective}

Suppose our objective is to maximize the average reward (or the discounted reward) obtained by the stochastic exploration policy $\beta_{\boldsymbol{\mu},\boldsymbol{\Sigma}}$ (shorted as $\beta$ when there is no ambiguity):
\begin{equation}
\label{eq:method-zoacobj-avg}
J^{\text{Avg}}(\boldsymbol{\mu},\boldsymbol{\Sigma})=\mathbb{E}_{s\sim d_{\beta_{\boldsymbol{\mu},\boldsymbol{\Sigma}}},a\sim\beta_{\boldsymbol{\mu},\boldsymbol{\Sigma}}}\left[r(s,a)\right],\quad \text{(Average)}
\end{equation}
\begin{equation}
\label{eq:method-zoacobj-dis}
J^{\text{Dis}}(\boldsymbol{\mu},\boldsymbol{\Sigma})=\mathbb{E}_{s\sim d_0}\left[v^{\beta_{\boldsymbol{\mu},\boldsymbol{\Sigma}}}(s)\right],\quad \text{(Discounted)}
\end{equation}
which can be seen as a Gaussian smoothed objective of the original sequential decision problem \eqref{eq:prel-objpi-avg} or \eqref{eq:prel-objpi-dis}. Since we are performing step-wise exploration in the parameter space, we further define $\tilde{\boldsymbol{\xi}}\in\mathbb{R}^{DT}$ as the concatenation of $D$-dimensional perturbations of a total of $T$ steps. For the purposes of analysis, we define $\tilde{f}:\mathbb{R}^{DT}\rightarrow\mathbb{R}$ as a function of the concatenated parameter vector and suppose the second-order derivative exists. The function in the local neighborhood of $\tilde{\boldsymbol{\mu}}\in\mathbb{R}^{DT}$ (concatenation of the same $\boldsymbol{\mu}$) can be approximated using the second-order Taylor expansion:
\vspace{-0.02in}
\begin{equation*}
\tilde{f}(\tilde{\boldsymbol{\mu}}+{\tilde{\boldsymbol{\Sigma}}}^{\frac{1}{2}}\tilde{\boldsymbol{\xi}})\approx \tilde{f}(\tilde{\boldsymbol{\mu}})+\tilde{\boldsymbol{g}}(\tilde{\boldsymbol{\mu}})^T{\tilde{\boldsymbol{\Sigma}}}^{\frac{1}{2}}\tilde{\boldsymbol{\xi}}+\frac{1}{2}\tilde{\boldsymbol{\xi}}^T{\tilde{\boldsymbol{\Sigma}}}^{\frac{1}{2}}\tilde{\boldsymbol{H}}(\tilde{\boldsymbol{\mu}}){\tilde{\boldsymbol{\Sigma}}}^{\frac{1}{2}}\tilde{\boldsymbol{\xi}},
\end{equation*}
where $\tilde{\boldsymbol{\Sigma}}$ is a block diagonal matrix, also concatenated from $\boldsymbol{\Sigma}$. If the value function $\tilde{f}$ is concave, then the Hessian matrix $\tilde{\boldsymbol{H}}$ is negative semi-definite. The Gaussian smoothed objective serves as a lower bound of the expected episodic return using policy $\pi_{\boldsymbol{\mu}}$ (supposing the Taylor expansion is exact):
\vspace{-0.02in}
\begin{align*}
J(\boldsymbol{\mu},\boldsymbol{\Sigma})&=\mathbb{E}_{\tilde{\boldsymbol{\xi}}\sim \mathcal{N}(\boldsymbol{0},\boldsymbol{I}_{DT})}\tilde{f}(\tilde{\boldsymbol{\mu}}+{\tilde{\boldsymbol{\Sigma}}}^{\frac{1}{2}}\tilde{\boldsymbol{\xi}})\\
&=\tilde{f}(\tilde{\boldsymbol{\mu}})+\frac{1}{2}\mathrm{tr}\left({\tilde{\boldsymbol{\Sigma}}}^{\frac{1}{2}}\tilde{\boldsymbol{H}}(\tilde{\boldsymbol{\mu}}){\tilde{\boldsymbol{\Sigma}}}^{\frac{1}{2}}\right)\\
&\leq\tilde{f}(\tilde{\boldsymbol{\mu}})=f(\boldsymbol{\mu})=\mathbb{E}_{s\sim d_0}[v^{\pi_{\boldsymbol{\mu}}}(s)].
\end{align*}
Intuitively, optimizing the Gaussian smoothed objective may encourage the agent to find robust areas of the search space (i.e., wide optima) than using the original objective \cite{lehman2018more}.

\subsubsection{Gradient Derivation}
Referencing to the proof of the policy gradient theorem \cite{sutton2000policy, sutton2018reinforcement} and using the ES-style gradient estimates \eqref{eq:prel-esgrad.mean}, we can derive the following theorem.

\begin{theorem}[Zeroth-order policy gradient]
For any Markov Decision Process, the zeroth-order policy gradient of the optimization objective \eqref{eq:method-zoacobj-avg} or \eqref{eq:method-zoacobj-dis} with respect to the parameters of the Gaussian distribution is
\begin{align}
\label{eq:method-zoacgrad.mean}
\!\!\!\!\!\nabla_{\boldsymbol{\mu}} J({\boldsymbol{\mu}},{\boldsymbol{\Sigma}})&=\mathbb{E}_{s\sim d_{\beta}}\mathbb{E}_{\boldsymbol{\theta}\sim\mathcal{N}(\boldsymbol{\mu},\boldsymbol{\Sigma})}[q^\beta(s,\pi_{\boldsymbol{\theta}}(s))\boldsymbol{w}_{\boldsymbol{\mu},\boldsymbol{\Sigma}}(\boldsymbol{\theta})],\!\!\!
\end{align}
\end{theorem}

\begin{proof}
We prove the theorem for discounted reward formulation \eqref{eq:method-zoacobj-dis} and defer the poof the theorem under average reward formulation \eqref{eq:method-zoacobj-avg} to the supplementary material. 

First, we can unroll $\nabla_{\boldsymbol{\mu}}v^\beta(s)$ step by step till infinity as:
\vspace{-0.02in}
\begin{align*}
&\nabla_{\boldsymbol{\mu}}v^\beta(s)\\
&=\nabla_{\boldsymbol{\mu}}\mathbb{E}_{\boldsymbol{\theta}\sim\mathcal{N}(\boldsymbol{\mu},\boldsymbol{\Sigma})}[q^\beta(s,\pi_{\boldsymbol{\theta}}(s))]\ \text{(shorted as $\mathbb{E}_{\boldsymbol{\theta}}[\cdot]$ in below)}\\
&=\mathbb{E}_{\boldsymbol{\theta}}[q^\beta(s,\pi_{\boldsymbol{\theta}}(s))\boldsymbol{w}_{\boldsymbol{\mu},\boldsymbol{\Sigma}}(\boldsymbol{\theta})]+\mathbb{E}_{\boldsymbol{\theta}}[\nabla_{\boldsymbol{\mu}}q^\beta(s,\pi_{\boldsymbol{\theta}}(s))]\\
&=\mathbb{E}_{\boldsymbol{\theta}}[q^\beta(s,\pi_{\boldsymbol{\theta}}(s))\boldsymbol{w}_{\boldsymbol{\mu},\boldsymbol{\Sigma}}(\boldsymbol{\theta})]\\
&\quad+\mathbb{E}_{\boldsymbol{\theta}}\Big[\nabla_{\boldsymbol{\mu}}\big(r(s,\pi_{\boldsymbol{\theta}}(s))+\int_\mathcal{S} \gamma\mathcal{P}_{s,s'}^{\pi_{\boldsymbol{\theta}}}v^\beta(s')\mathrm{d}s'\big)\Big]\\
&=\mathbb{E}_{\boldsymbol{\theta}}\left[q^\beta(s,\pi_{\boldsymbol{\theta}}(s))\boldsymbol{w}_{\boldsymbol{\mu},\boldsymbol{\Sigma}}(\boldsymbol{\theta})\right]+\mathbb{E}_{\boldsymbol{\theta}}\int_\mathcal{S}\gamma\mathcal{P}_{s,s'}^{\pi_{\boldsymbol{\theta}}}\nabla_{\boldsymbol{\mu}}v^\beta(s')\mathrm{d}s'\\
&=\mathbb{E}_{\boldsymbol{\theta}}\left[q^\beta(s,\pi_{\boldsymbol{\theta}}(s))\boldsymbol{w}_{\boldsymbol{\mu},\boldsymbol{\Sigma}}(\boldsymbol{\theta})\right]\\
&\qquad+\int_\mathcal{S}\gamma p(s\rightarrow s', 1, \beta)\mathbb{E}_{\boldsymbol{\theta}}\left[q^\beta(s',\pi_{\boldsymbol{\theta}}(s'))\boldsymbol{w}_{\boldsymbol{\mu},\boldsymbol{\Sigma}}(\boldsymbol{\theta})\right]\mathrm{d}s'\\
&\qquad+\int_\mathcal{S}\gamma^2 p(s \rightarrow s', 2, \beta)\nabla_{\boldsymbol{\mu}} v^\beta(s')\mathrm{d}s'\\
% &=\cdots
&=\int_\mathcal{S}\sum_{t=0}^{\infty}\gamma^t p(s\rightarrow s', t, \beta)\mathbb{E}_{\boldsymbol{\theta}}\left[q^\beta(s',\pi_{\boldsymbol{\theta}}(s'))\boldsymbol{w}_{\boldsymbol{\mu},\boldsymbol{\Sigma}}(\boldsymbol{\theta})\right]\mathrm{d}s'
\end{align*}
where $p(s\rightarrow s', t, \beta)$ is the probability of going from state $s$ to state $s'$ in $t$ steps under policy $\beta$. 
Note that we use the ES-style zeroth-order gradient \eqref{eq:prel-esgrad.mean} to get the third line of the derivation. 
By taking expectation over initial state distribution we can derive the zeroth-order policy gradient \eqref{eq:method-zoacgrad.mean}:
\begin{align*}
\nabla_{\boldsymbol{\mu}} J^{\text{Dis}}({\boldsymbol{\mu}},{\boldsymbol{\Sigma}})&=\nabla_\theta\mathbb{E}_{s\sim d_0}[v^{\beta}(s)]\\
&=\int_\mathcal{S}d_0(s)\nabla_{\boldsymbol{\mu}}v^{\beta}(s)\mathrm{d}s\\
&=\int_\mathcal{S}\int_\mathcal{S}\sum_{t=0}^{\infty}\gamma^t d_0(s)p(s\rightarrow s', t, \beta)\\
&\qquad\qquad\mathbb{E}_{\boldsymbol{\theta}}\left[q^\beta(s',\pi_{\boldsymbol{\theta}}(s'))\boldsymbol{w}_{\boldsymbol{\mu},\boldsymbol{\Sigma}}(\boldsymbol{\theta})\right]\mathrm{d}s'\mathrm{d}s\\
&=\mathbb{E}_{s\sim d_{\beta}}\mathbb{E}_{\boldsymbol{\theta}\sim\mathcal{N}(\boldsymbol{\mu},\boldsymbol{\Sigma})}[q^\beta(s,\pi_{\boldsymbol{\theta}}(s))\boldsymbol{w}_{\boldsymbol{\mu},\boldsymbol{\Sigma}}(\boldsymbol{\theta})].
\end{align*}
where the discounted state distribution $d_\beta$ under discounted reward formulation is defined in \eqref{eq:prel-distribution}.
\end{proof}
\vspace{-0.05in}
The zeroth-order policy gradient \eqref{eq:method-zoacgrad.mean} is inherently unbiased when optimizing the Gaussian smoothed objective \eqref{eq:method-zoacobj-avg} or \eqref{eq:method-zoacobj-dis}. The remaining problem is how to estimate the action-value term $q^\beta(s,\pi_{\boldsymbol{\theta}}(s))$ accurately. It is well known that discounted and average reward objectives become equivalent as $\gamma\rightarrow 1$ \cite{sutton2018reinforcement, zhang2021policy}, and discounted reward criterion is more widely used in literature. We only focus on the discounted reward objective in the remainder of this paper.

\subsubsection{Variance Reduction via Advantage Estimation}
In light of existing actor-critic RL algorithms \cite{mnih2016asynchronous, schulman2017proximal}, since $\boldsymbol{w}_{\boldsymbol{\mu},\boldsymbol{\Sigma}}(\boldsymbol{\theta})$
% and $\boldsymbol{W}_{\boldsymbol{\mu},\boldsymbol{\Sigma}}(\boldsymbol{\theta})$ 
is zero-centered and $v^\beta(s)$ is not correlated to any specific $\boldsymbol{\theta}$, we can subtract $v^\beta(s)$ as a baseline for variance reduction but still obtain unbiased policy gradient estimates:
\begin{align}
\label{eq:method-zoacgradadv.mean}
\nabla_{\boldsymbol{\mu}} J({\boldsymbol{\mu}},{\boldsymbol{\Sigma}})&=\mathbb{E}_{s\sim d_{\beta}}\mathbb{E}_{\boldsymbol{\theta}\sim\mathcal{N}(\boldsymbol{\mu},\boldsymbol{\Sigma})}[A^{\pi_{\boldsymbol{\theta}}}_{\boldsymbol{\mu},\boldsymbol{\Sigma}}(s)\boldsymbol{w}_{\boldsymbol{\mu},\boldsymbol{\Sigma}}(\boldsymbol{\theta})],
\end{align}
where the advantage function, intuitively interpreted as how much better it is using a specific policy $\pi_{\boldsymbol{\theta}}$ than using the exploration policy $\beta_{\boldsymbol{\mu},\boldsymbol{\Sigma}}$ with random Gaussian noise, gives:
\begin{equation}
\label{eq:method-adv}
A^{\pi_{\boldsymbol{\theta}}}_{\boldsymbol{\mu},\boldsymbol{\Sigma}}(s)=q^\beta(s,\pi_{\boldsymbol{\theta}}(s))-v^\beta(s).
\end{equation}
As described in Section \ref{sec:framework.exploration}, each individual is explored for $N$ steps, then we can use the Generalized Advantage Estimation (GAE) method \cite{schulman2015high} to compute \eqref{eq:method-adv}. GAE is the exponential weighted sum between one-step TD residual and multi-step TD residual, achieving a better bias-variance trade-off. Besides, we approximate the discounted state distribution $d_\beta$ and the expectation of Gaussian distribution $\mathbb{E}_{\boldsymbol{\theta}\sim\mathcal{N}(\boldsymbol{\mu},\boldsymbol{\Sigma})}[\cdot]$ by calculating the empirical mean of on-policy samples:
\begin{align}
\label{eq:method-zoacadvest.mean}
\nabla_{\boldsymbol{\mu}}{\hat{J}^{\text{(ZOAC)}}}&\approx\frac{1}{nH}\sum_{i=1}^{n}\sum_{j=0}^{H-1}f^{\text{(ZOAC)}}_{i,j}\boldsymbol{w}_{\boldsymbol{\mu},\boldsymbol{\Sigma}}(\boldsymbol{\theta}_{i,j}),
\end{align}
in which we use the same notation as in Section \ref{sec:framework.exploration}, i.e., $n$ trajectories are sampled, $nH$ individuals are evaluated for $N$ steps each. The \textit{fitness} values are estimated using GAE with an adjustable hyperparameter $\lambda\in[0,1]$:
\begin{equation}
\label{eq:method-zoacadvest.adv}
f^{\text{(ZOAC)}}_{i,j}\!=\!\hat{A}_N^{\pi_{\boldsymbol{\theta}_{i,j}}}\!\!=\!\!\sum_{k=0}^{N-1}\!(\gamma\lambda)^k(r_{i,jN+k}\!+\!\gamma v^\beta_{i,jN+k+1}\!-\!v^\beta_{i, jN+k}).
\end{equation}

Apparently, the proposed method can be regarded as a kind of ES method, in which the parameters are updated towards improving the \textit{fitness} values in a zeroth-order way. However, the highly noisy Monte Carlo return \eqref{eq:prel-esgradest.ret} used in standard ES gradient estimator \eqref{eq:prel-esgradest.mean} is replaced by the advantage function \eqref{eq:method-zoacadvest.adv} in our proposed proposed zeroth-order policy gradient \eqref{eq:method-zoacadvest.mean}, which efficiently utilize the sequential nature to handle temporal credit assignment. 

We derive the upper bounds of variance for these two kinds of gradient estimators in Theorem \ref{theorem.bound} following the variance analysis scheme in \cite{zhao2011analysis}.
%For analytic simplicity, we constrain the covariance matrix $\boldsymbol{\Sigma}$ to be diagonal. 
The variance of random vectors $\boldsymbol{g}$ is defined as $\mathbb{D}[\boldsymbol{g}]=\mathrm{tr}(\mathbb{E}[(\boldsymbol{g}-\mathbb{E}[\boldsymbol{g}])(\boldsymbol{g}-\mathbb{E}[\boldsymbol{g}])^T])$. 

\vspace{0.1in}

\begin{theorem}[Variance upper bounds of gradient estimators]
\label{theorem.bound}
We consider the following assumptions on the problem structure to guarantee a well-defined finite reward:

(A) Bounded reward per step, $\sup_{s\in\mathcal{S},a\in\mathcal{A}}\!|r(s,a)|\!\leq\!R\!<\!\infty$,

(B) Unbounded non-positive reward per step, while the limit of infinite-horizon return exists and is bounded as $-\infty<-R_G\leq \mathbb{E}_{\pi}[\sum_{i=0}^{\infty}\gamma^{i}r(s_{i},a_{i})|s_0]\leq0$  (e.g., LQR with stabilizing feedback gains).

We have the following variance upper bounds for ES gradient estimators \eqref{eq:prel-esgradest.mean} and ZOAC gradient estimators \eqref{eq:method-zoacadvest.mean} under Assumption (A):
\begin{equation}
\begin{aligned}
\label{eq:method-grad.mean.varboundA}
\mathbb{D}[\nabla_{\boldsymbol{\mu}}{\hat{J}^{\textnormal{(ES)}}}]&\leq\frac{R^2S_1\left(1-\gamma^{NH}\right)^2}{n(1-\gamma)^2},\\
\mathbb{D}[\nabla_{\boldsymbol{\mu}}\hat{J}^{\textnormal{(ZOAC)}}]&\leq\frac{4R^2S_1\left(1-(\gamma\lambda)^N\right)^2}{nH(1-\gamma\lambda)^2}.
\end{aligned}
\end{equation}
and under Assumption (B):
\begin{equation}
\begin{aligned}
\label{eq:method-grad.mean.varboundB}
\mathbb{D}[\nabla_{\boldsymbol{\mu}}{\hat{J}^{\textnormal{(ES)}}}]\leq\frac{R_G^2S_1}{n},\quad
\mathbb{D}[\nabla_{\boldsymbol{\mu}}\hat{J}^{\textnormal{(ZOAC)}}]\leq\frac{R_G^2S_1}{nH}.
\end{aligned}
\end{equation}

\end{theorem}

\begin{proof}
Using the property of second order raw moment of Gaussian distribution and the property of trace, we have
\begin{align*}
\mathbb{D}[f(\boldsymbol{\xi})\boldsymbol{w}_{\boldsymbol{\mu},\boldsymbol{\Sigma}}(\boldsymbol{\xi})]&\leq\mathrm{tr}(\mathbb{E}_{\boldsymbol{\xi}}[f(\boldsymbol{\xi})\boldsymbol{w}_{\boldsymbol{\mu},\boldsymbol{\Sigma}}(\boldsymbol{\xi})\boldsymbol{w}_{\boldsymbol{\mu},\boldsymbol{\Sigma}}(\boldsymbol{\xi})^Tf(\boldsymbol{\xi})])\\
&\leq f_U^2\mathrm{tr}\left(\mathbb{E}_{\boldsymbol{\xi}}[\boldsymbol{\Sigma}^{-\frac{1}{2}}\boldsymbol{\xi}\boldsymbol{\xi}^T\boldsymbol{\Sigma}^{-\frac{1}{2}}]\right)\\
&=f_U^2\mathrm{tr}(\boldsymbol{\Sigma}^{-1}),
\end{align*}
in which $\mathrm{tr}(\boldsymbol{\Sigma}^{-1})$ can be further denoted as $S_1$ for brevity, and $f_U$ denotes the maximum possible fitness values in ES \eqref{eq:prel-esgradest.ret} or ZOAC \eqref{eq:method-zoacadvest.adv}. Under Assumption (A), we have
\begin{equation*}
|f_U^{\textnormal{(ES)}}|\leq\sum_{t=0}^{NH}\gamma^{t}R=\frac{1-\gamma^{NH}}{1-\gamma}R,\ 
|f_U^{\textnormal{(ZOAC)}}|\leq\frac{1-(\gamma\lambda)^N}{1-\gamma\lambda}2R,
\end{equation*}
while under assumption (B) we can only leverage the bound on the total return over the entire trajectory, i.e., $|f_U^{\textnormal{(ES)}}|=|f_U^{\textnormal{(ZOAC)}}|\leq R_G$.

Finally, we can complete the proof noticing that variance is inversely proportional to the sample size of the estimators, i.e., $n$ in ES and $nH$ in ZOAC.
\end{proof}
\vspace{0.1in}

We can give an intuitive comparison of these variance bounds. In the case when $\gamma = 0.99$, $\lambda = 0.97$, and $NH=1000$ (both are integers), The ratio between them is:
\begin{equation*}
\label{eq:method-varcomp}
\begin{split}
\frac{\overline{\mathbb{D}}[\nabla_{\boldsymbol{\mu}}{\hat{J}^{\textnormal{(ES)}}}]}{\overline{\mathbb{D}}[\nabla_{\boldsymbol{\mu}}{\hat{J}^{\textnormal{(ZOAC)}}}]}&=\frac{H(1-\gamma^{NH})^2(1-\gamma\lambda)^2}{4(1-\gamma)^2(1-(\gamma\lambda)^N)^2}\\
&\approx\frac{3.9H}{(1-0.96^{1000/H})^2},\quad\text{Assumption (A)},\\
\text{or}&=H,\qquad\qquad\qquad\qquad\text{Assumption (B)},
\end{split}
\end{equation*}
which is monotonically increasing with respect to $H$ and larger than one for all $H\geq1$, indicating that using the proposed zeroth-order policy gradient with advantage estimation has great potential in reducing variance. 

\subsubsection{Convergence Property} Suppose we use the following iteration mechanism in ZOAC starting from a certain $\boldsymbol{\mu}_{0}$:
\begin{equation}
\label{eq:method-iterate}
\boldsymbol{\mu}_{k+1}=\boldsymbol{\mu}_{k}+\alpha_k\nabla_{\boldsymbol{\mu}}{\hat{J}^{\textnormal{(ZOAC)}}}, k\geq0,
\end{equation}
We give the convergence property of ZOAC as follows.

\begin{theorem}[Convergence of ZOAC] 
With the following assumptions: (1) $\nabla_{\boldsymbol{\mu}}J$ is $L$-Lipschitz continuous, (2) Step sizes $\{\alpha_k\}$ are positive scalars satisfying $\sum_k\alpha_k=\infty,\sum_k\alpha_k^2<\infty$, then $\{\boldsymbol{\mu}_{k}\}$ computed by \eqref{eq:method-iterate} converges to a local optimum where $\lim_{k\rightarrow \infty}\nabla_{\boldsymbol{\mu}} J(\boldsymbol{\mu}_k,{\boldsymbol{\Sigma}})=0$ with probability 1.
\end{theorem}
\begin{proof}
From Theorems 1 and 2, we know that \eqref{eq:method-zoacadvest.mean} is an unbiased gradient estimator with bounded variance of the smoothed optimization objective \eqref{eq:method-zoacobj-dis}. Additionally, based on the assumptions of Theorem 2, we can assert that the objective has a finite upper bound. Combined with the Lipschitz condition (1) and the step size schedule (2), all necessary conditions for the convergence of standard stochastic gradient descent (SGD) algorithms are met (except that we are performing maximization). Detailed proof can be found in \cite{patel2022global}.
\end{proof}

In practice, we may use function approximators (e.g., neural networks as in our practical algorithm described in Section \ref{sec:framwork.prac}) to replace $q^\beta(s, \pi_{\boldsymbol{\theta}}(s))$, which introduces significant complexity into the convergence analysis. Given the widely accepted belief that neural networks have universal approximation capabilities \cite{hornik1991approximation}, it is reasonable to assume that the critic is sufficiently accurate over the entire state space. A more detailed finite-time convergence analysis that considers value approximation error is left for future research.

\subsection{Practical Zeroth-order Actor-critic Algorithm}
\label{sec:framwork.prac}

Building upon the foundation established in Section \ref{sec:framework.zopg}, we introduce the practical Zeroth-Order Actor-Critic (ZOAC) algorithm. The overall framework is depicted in Figure \ref{fig:intro-zoac}, and the pseudo-code is summarized in Algorithm \ref{code:zoac}. 

\subsubsection{Overall Framework}
ZOAC utilizes a synchronous parallel computing architecture built on the open-source distributed framework Ray \cite{moritz2018ray}. The architecture comprises several components, including \textit{samplers}, \textit{evaluators}, \textit{replay buffers}, and the \textit{learner}.
In each iteration, the parallelized \textit{samplers} draw individuals from the current population, which is represented by a Gaussian distribution in the search space. The samplers then generate corresponding policies, interact with the environment to collect trajectories, and store individual transitions in the \textit{replay buffer}. In this section, we continue to use the notation in Section \ref{sec:framework.exploration} to represent transition data. 
Using these experiences, the ZOAC \textit{learner} alternates between two tasks for multiple epochs: first-order policy evaluation (PEV) and zeroth-order policy improvement (PIM). The PEV phase trains the critic network to estimate the state-value function of the exploration policy, while the PIM phase updates the population using the zeroth-order policy gradient.
The \textit{evaluator} component is responsible for periodically evaluating the learned policies and recording their performance evaluation metrics.

\begin{algorithm}[t]
\caption{Zeroth-Order Actor-Critic (ZOAC)}
\begin{algorithmic}[1]
\label{code:zoac}
\STATE \textbf{Initialize:} Gaussian distribution of the initial population $\mathcal{N}(\boldsymbol{\mu},\boldsymbol{\Sigma})$, initial critic network $V_w$, policy decoder $q_\psi^{-1}$
\FOR{each iteration}
\FOR{sampler $i=1,2,...,n$}
\FOR{$j=0,1,...,H-1$}
\STATE Sample $\boldsymbol{\theta}_{i,j}\sim \mathcal{N}(\boldsymbol{\mu},\boldsymbol{\Sigma})$
\STATE Run the reconstructed policy $\pi_{\boldsymbol{\theta}_{i,j}}$ for $N$ steps
\STATE Store the sampled individual $\boldsymbol{\theta}_{i,j}$ and $N$-step transition data to the replay buffer $\mathcal{B}$
\ENDFOR
\ENDFOR
%\FOR{epoch $m=0,1,2,...,M-1$}
\FOR{$i=1,2,\cdots,n$}
\STATE Compute TD residuals $\delta_{i,t}^{V_w}$ \eqref{eq:method-prac.td} and state-value targets $\hat{G}_{i,t}$ \eqref{eq:method-prac.tarest} at all timesteps $t\in\{0,1,\cdots,T\}$
\STATE Compute advantages (fitness values) $\hat{A}_N^{\pi_{\boldsymbol{\theta}_{i,j}}}$ \eqref{eq:method-prac.advest} for all sampled individuals $j\in\{0,1,\cdots,H-1\}$
\ENDFOR
\STATE Update critic network $V_w$ through SGD \eqref{eq:method-prac.pevgrad} with a mini-batch size $B$ on all data in $\mathcal{B}$ for $M$ epoches
\STATE Update the mean of Gaussian distribution $\boldsymbol{\mu}$ along the zeroth-order policy gradient \eqref{eq:method-zoacadvest.mean}% \eqref{eq:method-zoacis.cov} 
\ENDFOR
\end{algorithmic}
\end{algorithm}

\subsubsection{Real-valued Policy Encoding}
Encoding scheme plays a crucial role in evolutionary algorithms as it determines the representation of candidate solutions in the search space, which is typically tailored to the specific problem being addressed. Commonly used encoding schemes encompass binary encoding, real-valued encoding, and tree-based encoding \cite{katoch2021review}. 

When optimizing the parameters within the pre-defined policy structure, for example, switching thresholds in FSMs \cite{zhang2017finite}, controller gains \cite{he2023noisy, huang2021accelerated, rahayu2022particle, azar2019implementation, ren2021lqr, malik2019derivative}, weight matrices in MPC \cite{jain2021optimal, maass2021zeroth, mohammadi2020linear, maass2020tuning, chin2019re}, parameters of neural networks \cite{jain2021pixels, salimans2017evolution, such2017deep, song2021esenas, risi2019deep}. It is important to note that these parameters often have clear physical meanings (e.g., mass, length, time) and are typically constrained within a reasonable range. This range can be determined by leveraging prior knowledge and understanding of the system and may vary significantly across different parameters. As a result, perturbation of the same intensity (controlled by the covariance matrix $\boldsymbol{\Sigma}$, usually isotropic at the beginning) may lead to different outcomes in different dimensions. Since the exploration during learning relies on local random search in the parameter space, it is critical to ensure the behavioral diversity of the entire population from the beginning \cite{salimans2017evolution, mania2018simple}. 

Without loss of generality, we use a normalized real-valued search space $\Theta= [-1, 1]^D$, where $D$ is the number of design parameters. The scaling and shifting factors $\psi$ can be determined using the lower bound and upper bound of policy parameters. With this encoding scheme, the ZOAC learner performs evolution operations such as selection and mutation on the normalized real-valued parameter vectors $\boldsymbol{\theta}\in\Theta$. The samplers and evaluator respectively use policies $\pi_{\boldsymbol{\theta}}$ and $\pi_{\boldsymbol{\mu}}$ to interact with the environment. Both of them are reconstructed from the normalized real-valued parameter vectors $\boldsymbol{\theta}$ and $\boldsymbol{\mu}$ respectively using the decoder $q_\psi^{-1}$, which contains program structure information and the scaling and shifting factors.

\subsubsection{First-order Policy Evaluation}
Similar to on-policy actor-critic methods \cite{mnih2016asynchronous, schulman2017proximal}, the state-value function of the behavior policy $v^\beta(s)$ can be estimated by a jointly optimized critic network $V_w(s)$. In the $i$-th trajectory with length $T$, the state-value target $\hat{G}_{i,t}$ for each state $s_{i,t}$ is calculated as \cite{andrychowicz2021what}:
\begin{equation}
\label{eq:method-prac.tarest}
\hat{G}_{i,t}=V_w(s_{i,t})+\sum_{k=0}^{T-t-1}(\gamma\lambda)^k\delta_{i,t+k}^{V_w}.
\end{equation}
where $\lambda\in[0,1]$ is the GAE hyper-parameter controlling the trade-off between bias and variance \cite{schulman2015high}, and the one-step TD residual $\delta_{i,t}^{V_w}$ using critic network is calculated as:
\begin{equation}
\label{eq:method-prac.td}
\delta_{i,t}^{V_w}=r_{i,t}+\gamma V_w(s_{i,t+1})-V_w(s_{i,t})
\end{equation}
In each iteration, a total of $n \times N \times H$ states $s$ (assuming no early terminated episodes) and their corresponding state-value targets $\hat{G}$ are collected in buffer $\mathcal{B}$ and used to train the critic network (subscripts of $s, \hat{G}$ are omitted here). The objective of PEV is to minimize the mean squared error (MSE) loss between the network outputs and the state-value targets:
\begin{equation}
\label{eq:method-prac.pev}
J_{\text{critic}}(w)=\mathbb{E}_{(s,\hat{G})\sim\mathcal{B}}\left[\frac{1}{2}\left(V_w(s)-\hat{G}\right)^2\right].
\end{equation}
In practice, the critic network is updated for $M$ epochs. At the beginning of each epoch, we assign the collected transitions randomly to minibatches of size $B$ and recompute state-value targets using \eqref{eq:method-prac.tarest}. The semi-gradient, not backpropagated to the state value target $\hat{G}$, is used to update the critic network:
\begin{equation}
\label{eq:method-prac.pevgrad}
\nabla_{w}J_{\text{critic}}(w)=\mathbb{E}_{(s,\hat{G})\sim\mathcal{B}}\left[\left(V_w(s)-\hat{G}\right)\frac{\partial V_w(s)}{\partial w}\right].
\end{equation}

\subsubsection{Zeroth-order Policy Improvement}
In each iteration, in total $n\times H$ individuals are sampled and evaluated. For each of them, we estimate their advantages via GAE \cite{schulman2015high}:
\begin{equation}
\begin{aligned}
\label{eq:method-prac.advest}
f^{\text{(ZOAC)}}_{i,j}=\hat{A}_N^{\pi_{\boldsymbol{\theta}_{i,j}}}=\sum_{k=0}^{N-1}(\gamma\lambda)^k\delta_{i,jN+k}^{V_w}.
\end{aligned}
\end{equation}
Then we update the policy parameter vector $\boldsymbol{\theta}$ following the zeroth-order policy gradient \eqref{eq:method-zoacadvest.mean}.

\begin{figure}[!t]
\centering
\includegraphics[width=1.0\linewidth, keepaspectratio=true,trim=25 55 25 25,clip]{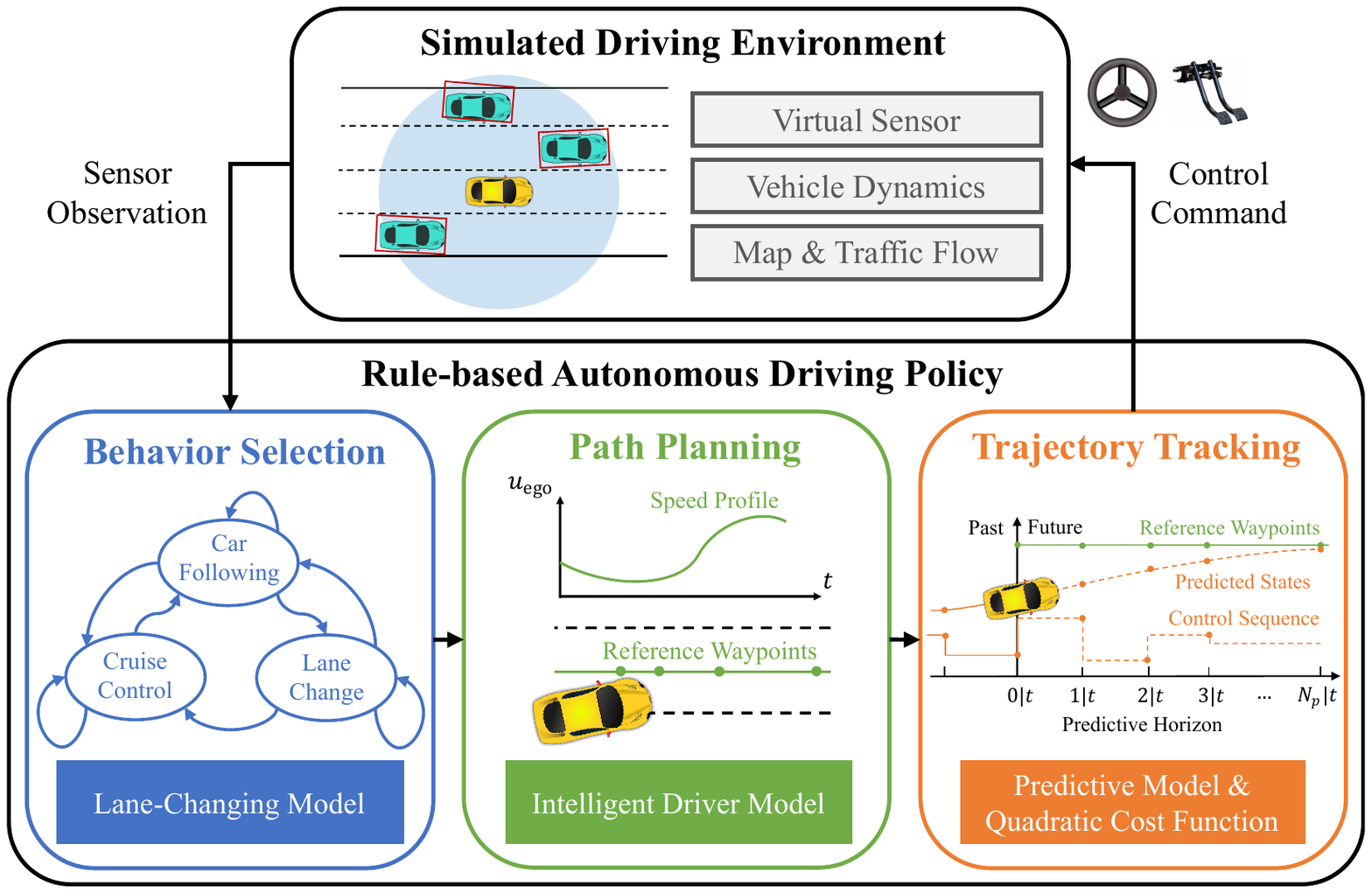}%
\vspace{-0.1in}
\caption{Task description of rule-based autonomous driving policy training.}
\vspace{-0.15in}
\label{fig:problem-adpolicy}
\end{figure}

\section{Tested Problem: Rule-based Autonomous Driving Policy Training}
\label{sec:problem}
Decision and control in autonomous driving is one of the main core functionalities of intelligent vehicles. Existing commercial autonomous driving systems typically adopt a hierarchical decision and control framework with multiple independently designed sub-modules and involve a large number of parameters to be adjusted. The reasonable selection of policy parameters is crucial to the driving performance, but heavily relies on considerable expertise of relevant field. To mitigate the time-consuming, laborious, and sub-optimal manual tuning process, automatic tuning methods that targeted for optimal driving performance are greatly desired. 
However, since the underlying optimization problem is high-dimensional, non-smooth, non-differentiable, and lacks closed-form expressions, evolutionary algorithms (EAs) are the dominating choice in existing literature \cite{likmeta2020combining, zhang2017finite}. 
Since the Zeroth-Order Actor-Critic (ZOAC) framework is an improvement over traditional EAs and should be especially suitable for solving this sequential decision problem, we use it as the primary testbed to verify the effectiveness of ZOAC. The rule-based autonomous driving policy training task is summarized in Figure \ref{fig:problem-adpolicy}, where the driving policy receives sensor observation from the simulated environment and outputs control command to the ego vehicle.

% \vspace{-0.1in}
\subsection{Simulated Driving Environment}
Compared to road test, simulated driving experiment has the advantages of low cost, high efficiency, and good reproducibility, making simulation platforms an indispensable part of the current development process of autonomous driving algorithms. In this section, we build a high-fidelity simulated driving environment for multi-lane scenarios, comprising virtual sensor module, vehicle dynamics module, map and traffic flow module \cite{guan2021learn, guan2022integrated, duan2024encoding, yang2023belief, ge2021numerically, krajzewicz2012recent, rakha2011virginia}. For more details about the environment, please refer to the supplementary material.

\begin{figure}[!t]
\centering
\includegraphics[width=0.8\linewidth, keepaspectratio=true,trim=48 420 75 400,clip]{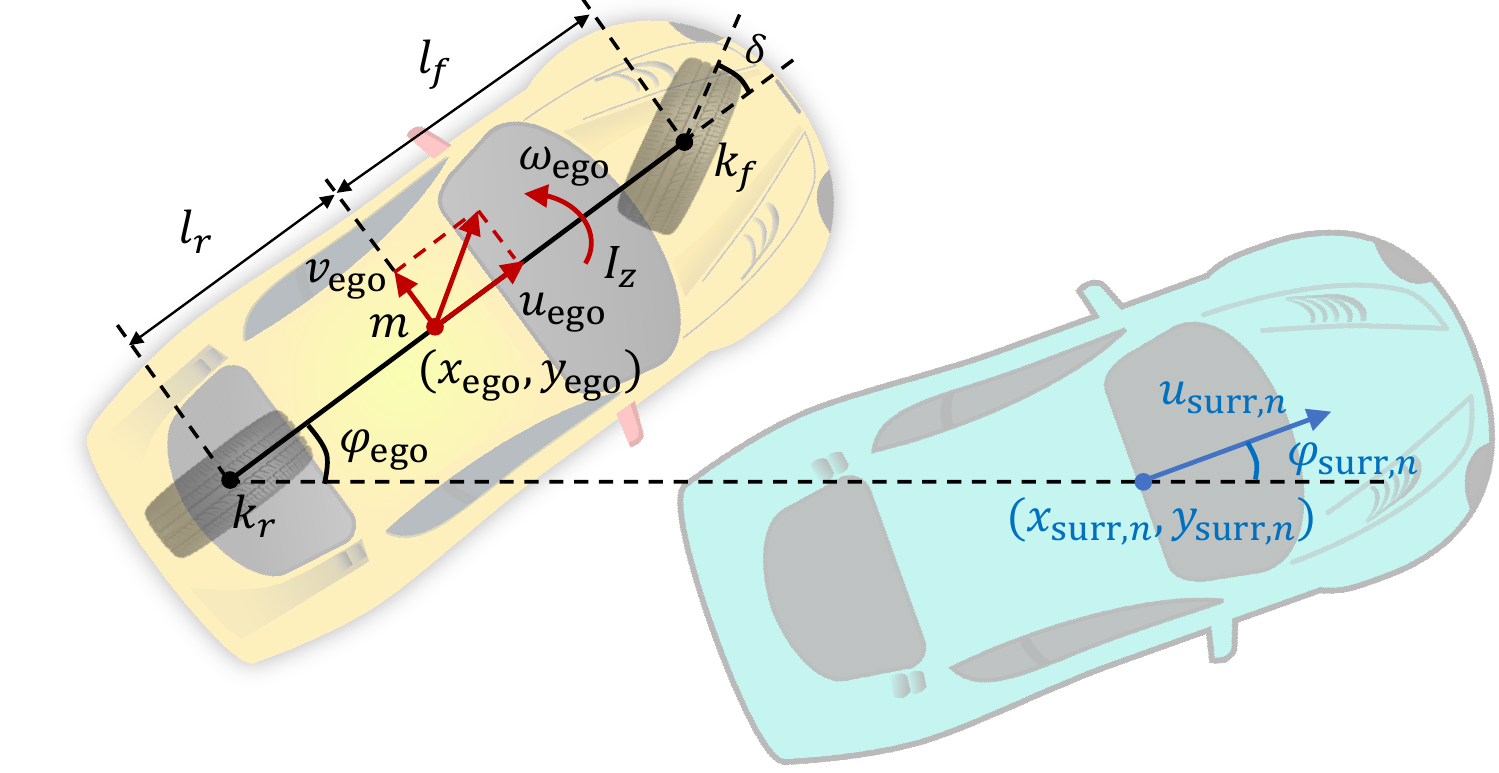}%
\vspace{-0.1in}
\caption{State of ego vehicle and surrounding vehicles.}
\vspace{-0.15in}
\label{fig:problem-egosurr}
\end{figure}

The target of ego vehicle is to drive as fast as possible under the premise of safety, comfort, economy, and compliance with traffic regulations. This process can be formulated as a Markov Decision Process (MDP) defined by a four-element tuple $(\mathcal{S}, \mathcal{A}, \mathcal{P}, r)$. At every step, the state $s$ includes the information of ego vehicle, surrounding vehicles and road geometry. Denote the number of surrounding vehicles within the detection range of ego vehicle as $n_d$. The state of ego vehicle and the state of the $n$-th detected surrounding vehicle can be represented as:
\begin{align}
\label{eq:problem-statedef}
\!\!\!s_{\text{ego}}&=[x_{\text{ego}}, y_{\text{ego}}, u_{\text{ego}}, v_{\text{ego}}, \varphi_{\text{ego}}, \omega_{\text{ego}}]^T,\\
\!\!\!s_{\text{surr},n}&=[x_{\text{surr},n}, y_{\text{surr},n}, \varphi_{\text{surr},n}, u_{\text{surr},n}]^T,\ n=1,\cdots,n_d,\!\!\!
\end{align}
where $x$ is horizontal position, $y$ is vertical position, $u$ is longitudinal velocity, $v$ is lateral velocity, $\varphi$ is yaw angle, and $\omega$ is yaw rate, as illustrated in Figure \ref{fig:problem-egosurr}. Road geometry information, obtained through map, is represented as several reference points including position and direction of lane centerlines ($y_{\text{center}}$ and $\varphi_{\text{center}}$).
The action $a$ is defined as the control commands of ego vehicle, which includes front wheel steering angle $\delta$ and acceleration command $a_x$:
\begin{equation}
\label{eq:problem-egoaction}
a=[\delta,a_x]^T,
\end{equation}
where we constrain $\delta\in[-0.3, 0.3]$ rad and $a_x\in[-5,2]$ m/s$^2$. 

The reward $r$ considers several aspects comprehensively, including driving efficiency, riding comfort, regulatory compliance, safety, and fuel economy \cite{guan2021learn, guan2022integrated}. We provide a detailed introduction of the reward function design in the supplementary material.

\subsection{Rule-based Autnomous Driving Policy}
We use a typical rule-based autonomous driving policy targeted for multi-lane scenarios, consisting of three sub-modules: behavior selection, path planning, and trajectory tracking. Each of them includes tunable design parameters that influence the overall driving performance, which are summarized in Table \ref{tab:adpolicypara}.
The upper layer of the policy selects driving behavior according to a finite-state machine (FSM), which includes three states: car following, cruise control, and lane change.
The middle and bottom layers are designed with reference to the Integrated Decision and Control (IDC) framework \cite{guan2021learn, guan2022integrated}. In the middle layer, the target lane and speed profile are determined based on current driving mode, and reference points are generated accordingly. The bottom layer controller is responsible for trajectory tracking under the model predictive control (MPC) framework.
Below we provide the detailed description of the autonomous driving policy. 

\begin{table}[t]
    \caption{Design parameters of the autonomous driving policy}
    \vspace{-0.05in}
    \label{tab:adpolicypara}
    \vspace{-0.1in}
    \begin{center}
    \begin{tabular}{lcc}
    \hline
    \textbf{Parameter}&\textbf{Unit}&\textbf{Range}\\
    \hline
    \emph{MOBIL Lane-changing Model} & & \\
    Politeness factor $p$ & / & [0, 1] \\
    Maximum safe deceleration $b_{\text{safe}}$ & m/s$^\text{2}$ & [0.5, 5.0] \\
    Lane-changing threshold $a_{\text{thres}}$ & m/s$^\text{2}$ & [0.05, 0.5] \\
    % Desired lane-changing time $t_{\text{lc}}$ & s & [1, 10] \\
    \hline
    \emph{Intelligent Driver Model}\\
    Maximum acceleration $a_{\text{max}}$ & m/s$^\text{2}$ & [0.5, 2.0] \\
    Desired deceleration $b_{\text{max}}$ & m/s$^\text{2}$ & [0.5, 5.0] \\
    Velocity exponent $\delta_u$ & / & [1, 10] \\
    Desired time headway $T$ & s & [1, 3] \\
    Jam distance $s_0$ & m & [1, 3] \\
    \hline
    \emph{Predictive Model of Ego Vehicle}\\
    Mass of the vehicle $m$ & kg & [1e3, 2e3]\\
    Yaw inertia of vehicle body $I_{z}$ & kg$\cdot$m$^\text{2}$ & [1e3, 2e3]\\
     Front axle sideslip stiffness $k_f$ & N/rad & [-16e4,-8e4] \\
     Rear axle sideslip stiffness $k_r$ & N/rad & [-16e4, -8e4]\\
     Distance from C.G. to front axle $l_f$ & m & [0.8, 2.2]\\
     Distance from C.G. to rear axle $l_r$ & m & [0.8, 2.2]\\
    \hline
    \emph{Quadratic Cost Function}\\
    Weighting factors of the stage cost $w_s^{\diamond}$ & / & [1e-5,1e3]\\
    Weighting factors of the terminal cost  $w_t^{\diamond}$ & / & [1e-5,1e3]\\
    \hline
    \end{tabular}
    \end{center}
    \vspace{-0.15in}
\end{table}

\subsubsection{Behavior Selection}
The rule-based behavior selection module is a finite-state machine (FSM) with three driving modes: cruise control, car following, and lane change. The switch conditions between driving modes mainly consider the surrounding vehicles in the current lane and adjacent lanes, as summarized in Table \ref{tab:switchfsm}.

\begin{table}[htb]
% \vspace{-0.1in}
    \caption{Mode switch conditions in behavior selection}
    \vspace{-0.15in}
    \label{tab:switchfsm}
    \begin{center}
    \begin{tabular}{lccc}
    \hline
    \textbf{Current state} & \textbf{Cruise control} & \textbf{Car Following} & \textbf{Lane Change} \\
    \hline
    Cruise control & Otherwise & \eqref{eq:problem-acc} & N/A\\
    Car Following & $\neg$\eqref{eq:problem-acc} & Otherwise & \eqref{eq:problem-mobil1}$\land$\eqref{eq:problem-mobil2}$\land$\eqref{eq:problem-lctime}\\
    Lane Change & $\neg$\eqref{eq:problem-acc}$\land$\eqref{eq:problem-lcend} & \eqref{eq:problem-acc}$\land$\eqref{eq:problem-lcend} & Otherwise \\
    \hline
    \end{tabular}
    \end{center}
    \vspace{-0.15in}
\end{table}

When no surrounding vehicle is in front of the ego vehicle, the FSM is in the cruise control mode, aiming to maintain a target cruise velocity. Otherwise, the ego vehicle follows the detected preceding car (denoted as CF). Therefore, the switch condition from cruise control mode to car following mode is
\begin{equation}
\label{eq:problem-acc}
x_{\text{CF}}-x_{\text{ego}}\leq x_{\text{thres}}.
\end{equation}

The MOBIL (Minimizing Overall Braking Induced by Lane change) model that considers both safety and lane-changing incentives is used to determine lateral behavior \cite{kesting2007general}. It ensures that the new follower after lane change does not suddenly decelerate and causes traffic accidents. Besides, the overall acceleration gain implies that the lane change is beneficial for the traffic flow. In addition, a minimum time interval between two consecutive lane changes is required to satisfy the traffic regulations. The switch condition from car following to lane changing can be summarized as:
\begin{align}
\label{eq:problem-mobil1}
\tilde{a}_{\text{new}}\geq&-b_{\text{safe}},\\
\label{eq:problem-mobil2}
\tilde{a}_{\text{ego}}-a_{\text{ego}}+p(\tilde{a}_{\text{new}}-a_{\text{new}}&+\tilde{a}_{\text{old}}-a_{\text{old}})\geq a_{\text{thres}},\\
\label{eq:problem-lctime}
\Delta t_{\text{lanekeep}}&>t_{\text{thres}},
\end{align}
where $a_{\text{ego}}$, $a_{\text{new}}$, $a_{\text{old}}$ respectively refers to the acceleration of ego vehicle, new follower and old follower before a prospective lane change, and $\tilde{a}_{\text{ego}}$, $\tilde{a}_{\text{new}}$, $\tilde{a}_{\text{old}}$ refers to the acceleration after a prospective lane change. When both adjacent lanes (left and right) are available, we simply choose the one that can bring larger acceleration gain. We leverage the Intelligent Driver Model (IDM) \cite{treiber2000congested} to compute the desired longitudinal acceleration based on relative velocity and time headway:
\begin{equation}
\begin{aligned}
\label{eq:problem-idm}
a_{\text{IDM}}&=a_{\mathrm{max}}\left[1-\left(\frac{u_{\text{ego}}}{u_{\text{target}}}\right)^{\delta_u}-\left(\frac{d^*}{\Delta d}\right)^2\right],\\
d^*&=s_0+u_{\text{ego}}T+\frac{u_{\text{ego}}\Delta u}{2\sqrt{a_{\mathrm{max}}b_{\mathrm{max}}}},
\end{aligned}
\end{equation}
where $\Delta d$ and $\Delta u$ is respectively the relative distance and velocity between the ego vehicle and the preceding vehicle, $d^*$ is the desired time headway, and $u_{\text{target}}$ is the target cruise velocity. Heuristic design parameters involved in \eqref{eq:problem-mobil1} \eqref{eq:problem-mobil2} \eqref{eq:problem-idm} are summarized in Table \ref{tab:adpolicypara}.

When the lateral position of ego vehicle approaches the centerline of the target lane with a deviation tolerance $y_{\text{thres}}$, the lane-changing process is considered complete and the driving mode is switched to either cruise control or car following: 
\begin{equation}
\label{eq:problem-lcend}
\vert y_{\text{ego}}-y_{\text{center}} \vert < y_{\text{thres}}.
\end{equation}

\subsubsection{Path Planning}
Based on the chosen driving behavior, we decompose the path planning problem into longitudinal and lateral planning, ultimately generating reference points as:
\begin{equation}
\!\!\!X^{\text{ref}}_{k|t}=[x^{\text{ref}}_{k|t}, y^{\text{ref}}_{k|t}, u^{\text{ref}}_{k|t}, v^{\text{ref}}_{k|t}, \varphi^{\text{ref}}_{k|t}, \omega^{\text{ref}}_{k|t}]^T\!,\ k=1,\cdots,N_p,\!\!
\end{equation}
where the subscript $\cdot_{k|t}$ denotes the state in the virtual predictive time step $k$ starting from the current time step $t$, $N_p$ is the predictive horizon. This notation will be used again later in the trajectory tracking module.

When the ego vehicle is under cruise control mode, the reference longitudinal velocity is the target cruise velocity $u_{\text{target}}$. Otherwise, the ego vehicle choose to follow the preceding vehicle using IDM \eqref{eq:problem-idm} with the parameters in Table \ref{tab:adpolicypara} and under the assumption that the preceding vehicle drives at a constant velocity. Then, based on the estimated reference acceleration, we can compute the reference longitudinal velocity $u^{\text{ref}}_{k|t}$ and position $x^{\text{ref}}_{k|t}$ of ego vehicle using forward Euler integration.
As for the lateral planning, the ego vehicle always aim to travel along the centerline of the target lane. Therefore, the reference lateral velocity $v^{\text{ref}}_{k|t}$ and yaw rate $\omega^{\text{ref}}_{k|t}$ should be zero, and the reference lateral position $y^{\text{ref}}_{k|t}$ and heading angle $\varphi^{\text{ref}}_{k|t}$ is equal to the position $y_{\text{center}}$ and direction $\varphi_{\text{center}}$ of the target lane centerline respectively. 
% On the other hand, during lane change, 
% we use a cubic polynomial curve to describe the lateral movement of ego vehicle \cite{mu2023neural}:
% \begin{equation}
% \label{eq:problem-poly}
% y(\tau)=y_0+y_1\tau+y_2\tau^2+y_3\tau^3,
% \end{equation}
% where $\tau\in[0,t_{\text{lc}}]$ denotes the time in lane change. Suppose the lateral position of current lane centerline is $y_a$ and that of the target lane centerline is $y_b$, the coefficients must ensure the continuity of reference trajectory in both position and velocity:
% \begin{equation*}
% y(0)=y_a,\ y(t_{\text{lc}})=y_b,\ \dot{y}(0)=\dot{y}(t_{\text{lc}})=0.
% \end{equation*}
% The polynomial coefficients in \eqref{eq:problem-poly} can then be solved as:
% \begin{equation}
% \!\!y_0=y_a,\ y_1=0,\ y_2=\frac{3(y_b-y_a)}{t_{\text{lc}}^2},\ y_3=-\frac{2(y_b-y_a)}{t_{\text{lc}}^3}.\!\!
% \end{equation}
% During lane change, the reference lateral position $y^{\text{ref}}_{k|t}$ and velocity $v^{\text{ref}}_{k|t}$ can be computed using the polynomial curve. Besides, the reference heading angle $\varphi^{\text{ref}}_{k|t}$ can be determined by calculating the ratio between longitudinal and lateral velocity, and the reference yaw rate remains zero.

% Note that the above path planning method does not involve online search or optimization process and therefore has high computational efficiency. All involved parameters are summarized in Table \ref{tab:adpolicypara}.

\subsubsection{Trajectory Tracking}
The bottom-level controller aims to track the reference points generated by planning module while avoiding collision. We adopt the model predictive control (MPC) framework to achieve this goal. At each time step in the real domain, a finite-horizon constrained optimal control problem is constructed: the primary objective is to closely track the $N_p$ reference points, using bicycle vehicle dynamics as the predictive model, and satisfying the amplitude constraints. This optimization problem can be mathematically represented as:
\begin{align*}
\min_{X,U} \sum_{k=0}^{N_p-1} &\gamma^k\!\left[(X_{k|t}-X^{\text{ref}}_{k|t})^T Q_{s}(X_{k|t}-X^{\text{ref}}_{k|t})+U_{k|t}^T R_sU_{k|t}\right]\!
\end{align*}
\vspace{-0.15in}
\begin{align}
\label{eq:problem-mpcobj}
+\gamma^{N_p}(X_{N_p|t}-X^{\text{ref}}_{N_p|t})^T Q_{t}(X_{N_p|t}-X^{\text{ref}}_{N_p|t}),
\end{align}
\vspace{-0.25in}
\begin{align} 
\label{eq:problem-mpc.ego}
\text{s.t.} \qquad \qquad \qquad \qquad X_{0|t}&=s_{\text{ego}}(t), \qquad \qquad \qquad \\ \label{eq:problem-mpc.egopred}
 X_{k+1|t}&=F_{\text{ego}}(X_{k|t},U_{k|t}),\\ \label{eq:problem-mpc.statecon}
X_{\mathrm{min}}\leq X_{k|t}&\leq X_{\mathrm{max}},\\ \label{eq:problem-mpc.actioncon}
U_{\mathrm{min}}\leq U_{k|t}&\leq U_{\mathrm{max}},
% d^{\text{v2v}}_{n,i,j}(k)&\geq D_{\text{safe}}^{\text{v2v}},\quad i,j\in\{1,2\},\\ \label{eq:problem-mpc.v2rcon}
% d^{\text{v2r}}_{i,j}(k)&\geq D_{\text{safe}}^{\text{v2r}},\quad i,j\in\{1,2\},
\end{align}
where the first state in virtual domain should be equal to the current state in real domain \eqref{eq:problem-mpc.ego}. The predictive horizon $N_p$ is set to 25 with a step size of 0.1s. The problem is repeatedly solved in an online manner and only the first action in the result will be applied as control input, i.e., $a(t)=U^*_{0|t}$.

The predicted states in virtual domain are encouraged to keep close to the reference points $X^{\text{ref}}_{k|t}$ by minimizing the quadratic objective function \eqref{eq:problem-mpcobj} with diagonal positive-definite weighting matrices:
\begin{align}
\label{eq:problem-wss}
Q_{s}&=\mathrm{diag}\left(w_s^x,w_s^y,w_s^u,w_s^v,w_s^\varphi,w_s^\omega\right),\\ \label{eq:problem-wst}
Q_{t}&=\mathrm{diag}\left(w_t^x,w_t^y,w_t^u,w_t^v,w_t^\varphi,w_t^\omega\right),\\ \label{eq:problem-was}
R_s&=\mathrm{diag}\left(w_s^\delta,w_s^{a_x}\right).
\end{align}
Note that the quadratic objective is widely used in MPC because of its high efficiency. However, the driving performance of true concern usually lacks a clear expression and needs to be approximated, e.g., the reward $r$ in the simulated environment whose analytic expression is supposed to be unknown when designing the controller. Therefore, a common practice in the MPC community is to elaborately adjust the weighting matrices to balance the cost of different state variables and control variables.

$F_{\text{ego}}$ represents the predictive model for ego vehicle \eqref{eq:problem-mpc.egopred}, a numerically stable dynamic bicycle model with linear tire model \cite{ge2021numerically} illustrated in Figure \ref{fig:problem-egosurr} and represented as:
\begin{equation}
    \!\!\!\!\left[\!\!\begin{array}{c}
    x_{k+1} \\
    y_{k+1} \\
    u_{k+1} \\
    v_{k+1} \\
    \varphi_{k+1} \\
    \omega_{k+1}
    \end{array}\!\!\right]\!\!=\!\!\left[\!\!\!\begin{array}{c}
    x_{k}+T_{s}\left(u_{k} \cos \varphi_{k}-v_{k} \sin \varphi_{k}\right) \\
    y_{k}+T_{s}\left(v_{k} \cos \varphi_{k}+u_{k} \sin \varphi_{k}\right) \\
    u_{k}+T_{s} a_{k} \\
    \frac{m u_{k} v_{k}+T_{s}\left(l_{f}k_{f}-l_{r} k_{r}\right) \omega_{k}-T_{s} k_{f} \delta_{k} u_{k}-T_{s} m u_{k}^{2} \omega_{k}}{m u_{k}-T_{s}\left(k_{f}+k_{r}\right)} \\
    \varphi_{k}+T_{s} \omega_{k} \\
    \frac{I_{z} u_{k} \omega_{k}+T_{s}\left(l_{f} k_{f}-l_{r} k_{r}\right) v_{k}-T_{s} l_{f} k_{f} \delta_{k} u_{k}}{I_{z} u_{k}-T_{s}\left(l_{f}^{2} k_{f}+l_{r}^{2} k_{r}\right)}
    \end{array}\!\!\!\right]\!\!,\!\!
    \label{eq:problem-egomodel}
\end{equation}
% \begin{equation}
%     \left[\!\!\begin{array}{c}
%     x^n_{k+1} \\
%     y^n_{k+1} \\
%     \varphi^n_{k+1} \\
%     u^n_{k+1}
%     \end{array}\!\!\right]\!\!=\!\!\left[\!\!\begin{array}{c}
%     x^n_{k}+T_{s}u^n_{k} \cos \varphi_{k} \\
%     y^n_{k}+T_{s}u^n_{k} \sin \varphi_{k} \\
%     \varphi^n_{k} \\
%     u^n_{k}
%     \end{array}\!\!\right]\!\!,
%     \label{eq:problem-surrmodel}
% \end{equation}
where $T_s$ is the step length in virtual horizon, and the subscripts $\cdot_{k|t}$ are shorted as $\cdot_{k}$ for brevity. According to the Identification for Control (I4C) paradigm \cite{gevers2005identification}, model learning can also be regarded as a controller design problem aiming to maximize control performance. In the trajectory tracking module, all weighting factors in the quadratic cost \eqref{eq:problem-wss} \eqref{eq:problem-wst} \eqref{eq:problem-was} and parameters involved in the ego vehicle predictive model \eqref{eq:problem-egomodel} are trained towards optimal driving performance, as summarized in Table \ref{tab:adpolicypara}.

The amplitude constraints \eqref{eq:problem-mpc.statecon} \eqref{eq:problem-mpc.actioncon} primarily consider the response capability of the vehicle chassis, stability, and compliance with traffic regulations. They construct box constraints for state and action, with $X_\mathrm{min}$, $U_\mathrm{min}$ be the minimum allowed values and $X_\mathrm{max}$, $U_\mathrm{max}$ be the maximum allowed values.

In practical implementation, the MPC controller leverages the open-source numerical optimization framework CasADi, and chooses the ipopt solver based on the primal-dual interior point method to solve the constrained optimizaiton problem \eqref{eq:problem-mpcobj} in an online way \cite{andersson2019casadi}.

\section{Experiments and Discussions}
\label{sec:exp}

\subsection{Experimental Setup}
\label{sec:exp-setup}

In this section, we aim to compare the proposed ZOAC algorithm with popular zeroth-order optimization baselines in solving sequential decision problems. We leverage the open-source gradient-free optimization library NeverGrad \cite{nevergrad} and follow the default hyperparameter settings. Specifically, we use the following three baselines:

(1) Covariance Matrix Adaptation Evolutionary Strategy (CMA-ES) \cite{hansen2001completely, jain2021optimal, o2020tunercar}, a popular variant of ES with covariance matrix and step size adaptation. We use the default setting where the population size is $\max\{n,4+\left\lfloor3\ln{D}\right\rfloor\}$.

(2) Particle Swarm Optimization (PSO) \cite{kennedy1995particle, rahayu2022particle, azar2020implementation, xu2019automated} that moves particles around according to each particle's current position and velocity, as well as the best-known positions in the search space found by the swarm. Hyperparameters include population size, inertia weight $\omega$, cognitive coefficient $\phi_p$ and social coefficient $\phi_g$, all of which are set to their default values.

(3) Differential Evolution (DE) \cite{storn1997differential, sahu2021differential} that uses differences between points in the population to mutate parameters in promising directions. We use a two-points crossover variant and set the differential weights $F_1, F_2$ to 0.8 as recommended.

\begin{table}[b]
\vspace{-0.1in}
    \caption{Hyper-parameters of ZOAC in the autonomous driving task}
    \label{tab:hyper}
    \vspace{-0.1in}
    \begin{center}
    \begin{tabular}{lc}
    \hline
    \textbf{Hyper-parameter}&\textbf{Value}\\
    \hline
    Number of samplers $n$ & 8 \\
    Segment length $N$ & 40 \\
    Train frequency $H$ & 25 \\
    Noise standard deviation $\sigma$ & 0.1 \\
    Discount factor $\gamma$ & 0.99 \\
    GAE coefficient $\lambda$ & 0.97 \\
    Critic update epoch $M$ & 5 \\
    Minibatch size $B$ & 128 \\
    Critic network $V_w$ & (256, 256) w/ tanh\\
    Critic optimizer & AdamW($\beta_1$ = 0.9, $\beta_2$ = 0.999)\\
    Critic learning rate $\alpha_\text{critic}$ & $\text{5e-4}$\\
    Actor learning rate $\alpha_\text{actor}$ & $\text{5e-4}$\\
    \hline
    \end{tabular}
    \end{center}
    %\vspace{-0.1in}
\end{table}

\begin{table*}[b]
\centering
\vspace{-0.1in}
\caption{Comparison of driving performance.}
\label{tab:comp}
\vspace{-0.1in}
\begin{center}
\begin{tabular}{c|cccc|c}
\hline
\textbf{Method} & \textbf{ZOAC} & \textbf{CMA-ES} & \textbf{PSO} & \textbf{TwoPointDE} & \textbf{Initial}\\ 
\hline
Total Average Return $\uparrow$ & 5139.25$\pm$891.75  & -159.54$\pm$3999.29  & -389.03$\pm$7545.62 & 3822.80$\pm$2618.67 & -2483.90$\pm$4174.44 \\
Average Speed (km/h) $\uparrow$ & 41.45$\pm$2.57  & 41.66$\pm$2.60  & 41.67$\pm$3.20 & 41.20$\pm$3.05 & 41.89$\pm$2.83 \\
Riding Comfort (m/s$^{\text{2}}$) $\downarrow$ & 0.38$\pm$0.19 & 1.18$\pm$0.49 & 1.17$\pm$1.08 & 0.60$\pm$0.57 & 1.35$\pm$0.45 \\
Fuel Efficiency \cite{rakha2011virginia} (L/100km) $\downarrow$ & 5.09$\pm$0.44 & 5.11$\pm$0.33 & 5.27$\pm$1.00 & 5.09$\pm$0.59 & 5.20$\pm$0.35 \\
% Training Time (s) $\downarrow$ & 688.30$\pm$8.03  & 391.43$\pm$6.13  & 473.59$\pm$24.68 & 429.59$\pm$18.71 & / \\
\hline
\end{tabular}
\end{center}
\end{table*}

The hyperparameters of ZOAC used in the experiments are summarized in Table \ref{tab:hyper}. We will justify our choices through a detailed ablation study on hyperparameters in Section \ref{sec:exp-ablation}. All experiments in this paper were conducted on a Macbook Pro with an Apple M2 Max CPU and 32 GB RAM, with each experiment run 10 times using different random seeds.

\subsection{Main Results in Autonomous Driving Task}

\subsubsection{Performance Evaluation}
\label{sec:exp-performance}

\begin{figure}[!t]
\centering
\subfloat[]{\includegraphics[height=1.9in, keepaspectratio=true,trim=50 300 50 48,clip]{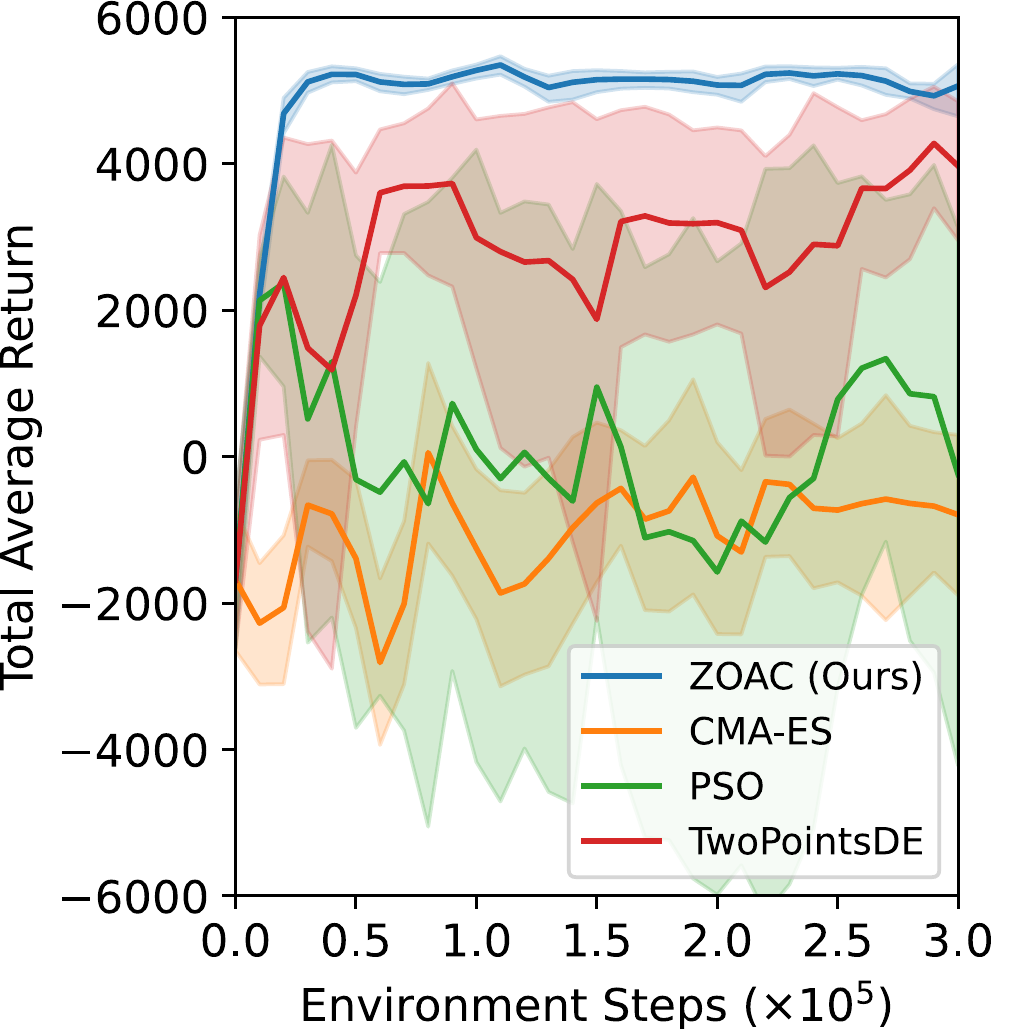}%
\label{fig:exp-main_result-avg}}
% \vspace{0.01in}
\subfloat[]{\includegraphics[height=1.9in, keepaspectratio=true,trim=30 300 160 38,clip]{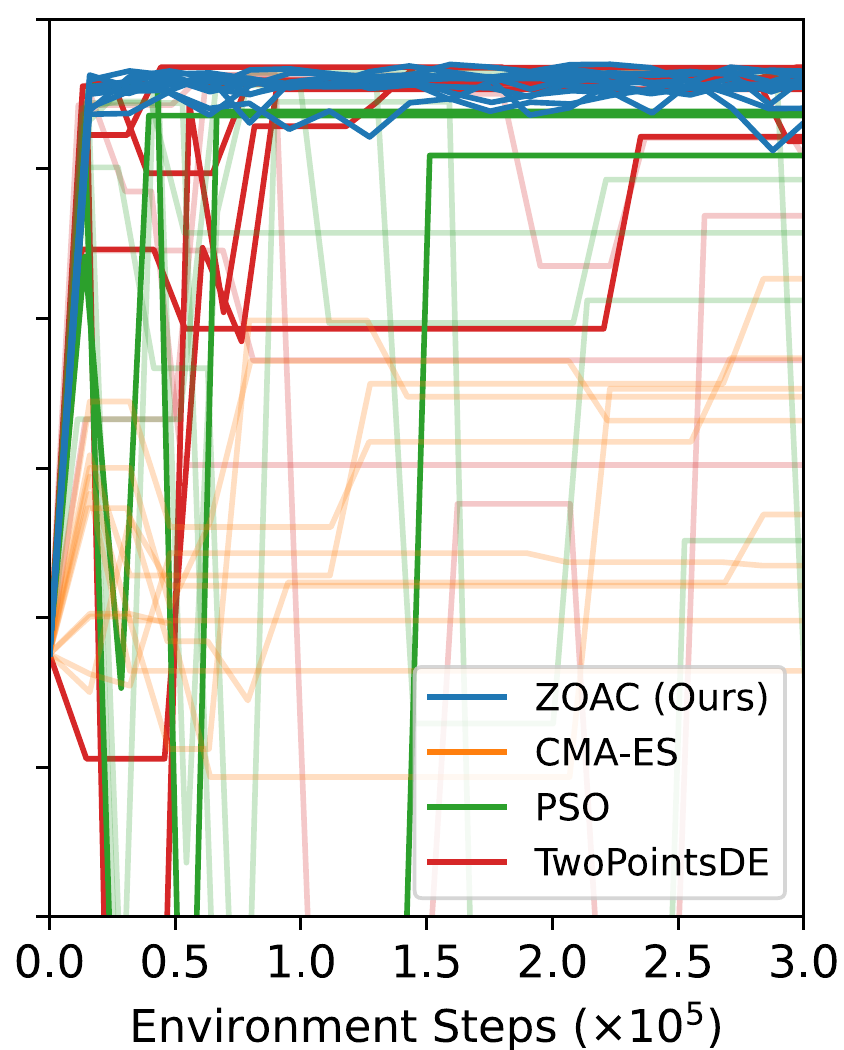}%
\label{fig:exp-main_result-perseed}}
\caption{Learning curves of ZOAC and baseline methods in the autonomous driving task: (a) The solid lines represent the mean, and the shaded regions indicate the 95\% confidence interval over 10 random seeds. All learning curves in this paper are presented in this manner unless otherwise specified. (b) Each solid line represents an independent training run, with runs that have a final performance not exceeding 4000 set to translucent for visual clarity.}
\vspace{-0.15in}
\label{fig:exp-main_result}
\end{figure}

The autonomous driving policy is evaluated every two iterations, with each evaluation reporting the total average return over 8 episodes using the recommended parameter values. Specifically, for ZOAC, we use the mean $\boldsymbol{\mu}$ of the current population. For CMA-ES, PSO, and TwoPointsDE, we use the \texttt{recommend()} API provided by NeverGrad to obtain the historically best candidates. These parameters will be saved as checkpoints for further evaluation. However, due to the stochastic nature of the multi-lane driving environment, the driving performance of the recommended policies does not necessarily exhibit a monotonic increase (as in the deterministic case) throughout the learning process. To clearly illustrate this process, in Figure \ref{fig:exp-main_result-perseed}, we directly use the previous evaluation results when the policy remains unchanged, resulting in horizontal lines upon convergence. 

Figure \ref{fig:exp-main_result-avg} shows that the proposed ZOAC algorithm significantly outperforms baselines in terms of sample efficiency, training stability, and final performance. Figure \ref{fig:exp-main_result-perseed} also highlights the advantage of ZOAC by showing that all 10 runs find a policy that can achieve an episodic return of over 4,000, compared to 6 out of 10 for TwoPointsDE, 3 out of 10 for PSO, and none for CMA-ES.

Table \ref{tab:comp} reported the final driving performance (mean $\pm$ standard deviation) of the learned policies. Specifically, we use the checkpoint saved at 3e5 timesteps as final policies, then evaluate them in the multi-lane scenario for 100 episodes with a duration of 100 seconds. The riding comfort $I_{\text{comfort}}$ is calculated as the root mean square (RMS) of longitudinal and lateral acceleration:
\begin{equation*}
\begin{aligned}
I_{\text{comfort}}=1.4\sqrt{a_{x,\text{RMS}}^2+a_{y,\text{RMS}}^2},\quad a_{\cdot,\text{RMS}}=\sqrt{\frac{1}{N}\sum_{i=1}^{N}{a_{\cdot,i}^2}}.
\end{aligned}
\end{equation*}
where $a_{x,i}$ and $a_{y,i}$ represent the lateral and longitudinal acceleration at the $i$-th step of episode, respectively, $N$ is the number of data points. The fuel efficiency is estimated by Virginia Tech comprehensive power-based fuel consumption model (VT-CPFM) \cite{rakha2011virginia} over episodes.

\begin{figure}[!t]
\centering
\subfloat[]{\includegraphics[height=1.9in, keepaspectratio=true,trim=50 280 48 48,clip]{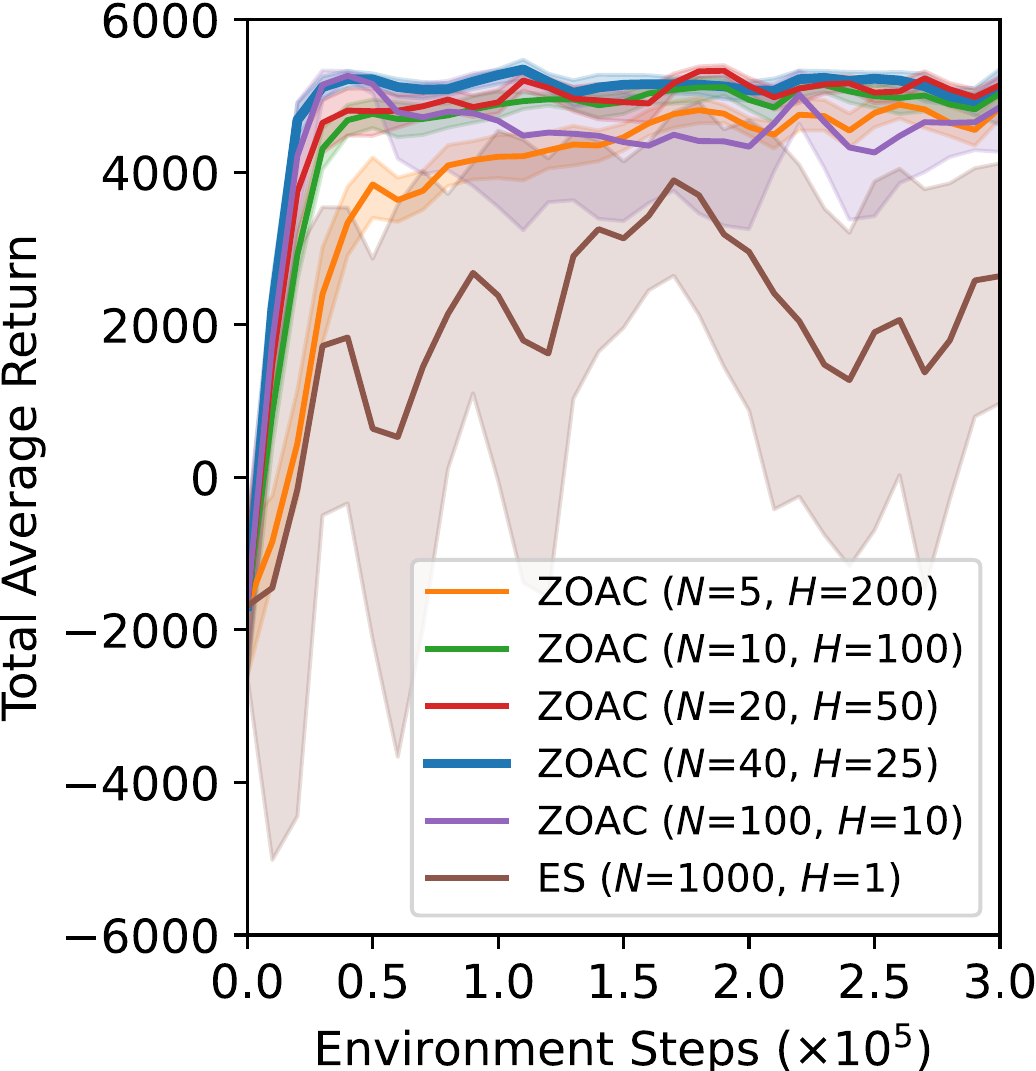}%
\label{fig:exp-ablation-NH}}
% \vspace{0.01in}
\subfloat[]{\includegraphics[height=1.9in, keepaspectratio=true,trim=30 301 160 37,clip]{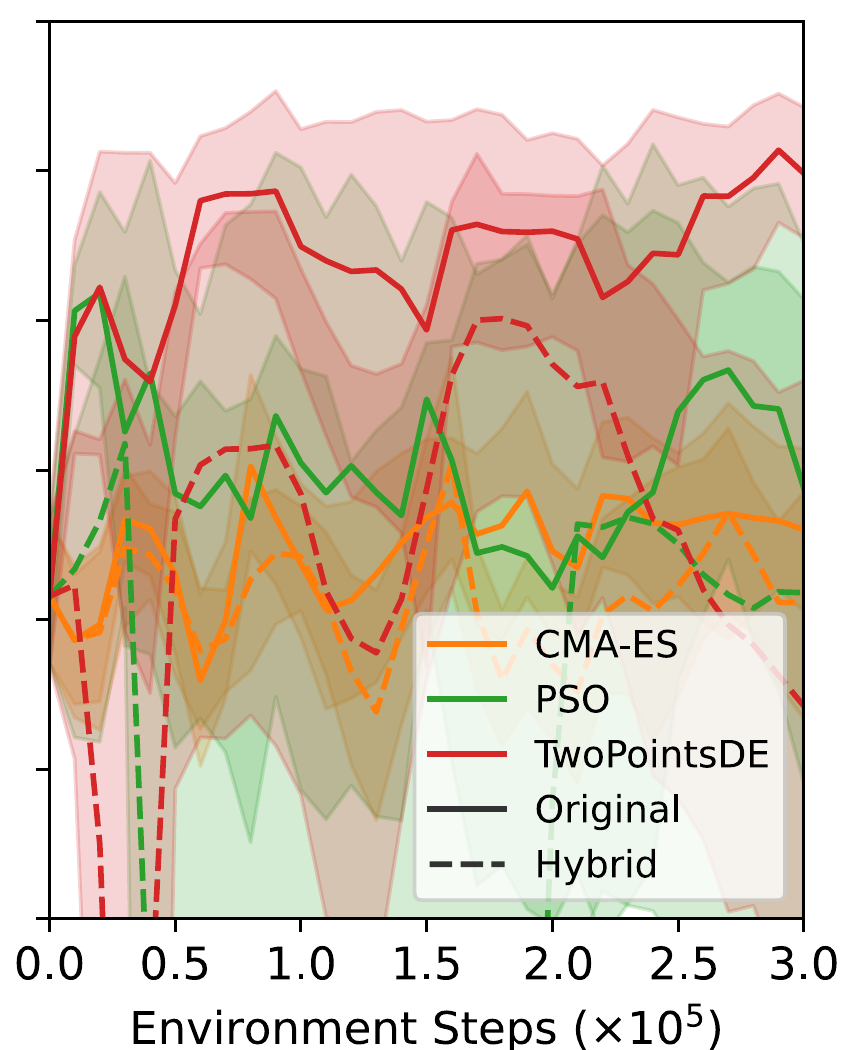}%
\label{fig:exp-ablation-hybridbaseline}}
\vspace{0.01in}
\subfloat[]{\includegraphics[height=0.9in, keepaspectratio=true,trim=40 655 20 38,clip]{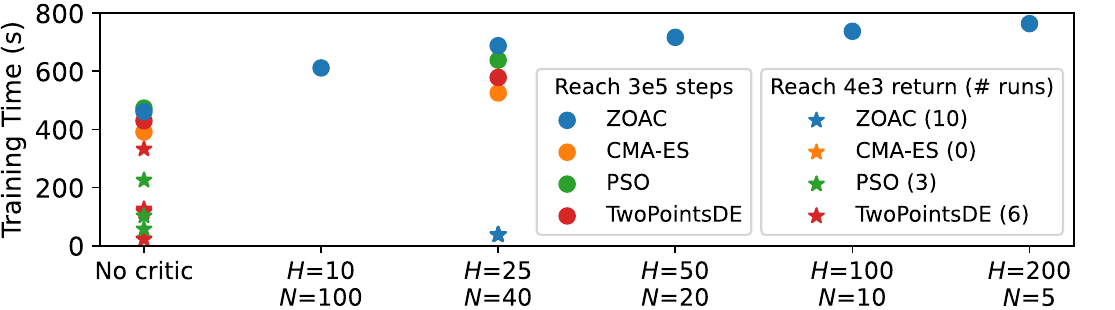}%
\label{fig:exp-ablation-time}}
\caption{Ablation study on the hybridization of actor-critic architecture. (a) Learning curves of ES and ZOAC with different segment length $N$ and training frequency $H$. (b) Heuristic hybridization of baselines and critic. Solid lines represent the original baselines, while dashed lines represent the hybrid ones. We use $N=40$ and $H=25$ for all three baselines. (c) Statistics of the wall-clock training time. Circles indicate the average total training time for 3e5 environment steps over 10 independent runs, while stars represent the time when the policy consistently exceeds a return of 4,000 in a single run. The number indicates how many out of the 10 runs achieved this milestone.}
\vspace{-0.15in}
\label{fig:exp-ablation_hybrid}
\end{figure}

\begin{figure*}[t]
\centering
\subfloat[Parameter evolution curves]{\includegraphics[width=0.27\linewidth, keepaspectratio=true,trim=30 620 270 30,clip]{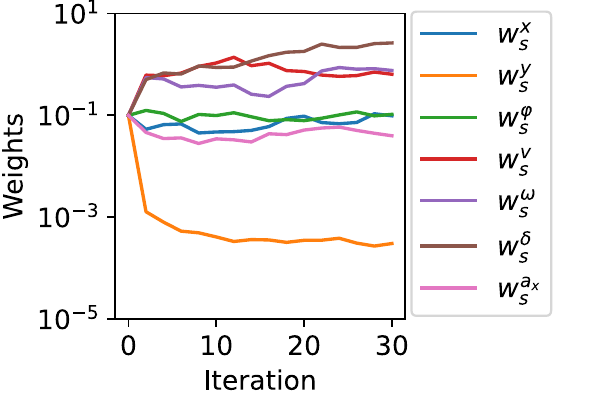}%
\label{fig:exp-mean_evo}}
%\hspace{-0.1in}
\subfloat[State and action variables of a typical\\lane-changing trajectory (Before training)]{\includegraphics[width=0.35\linewidth, keepaspectratio=true,trim=45 560 60 40,clip]{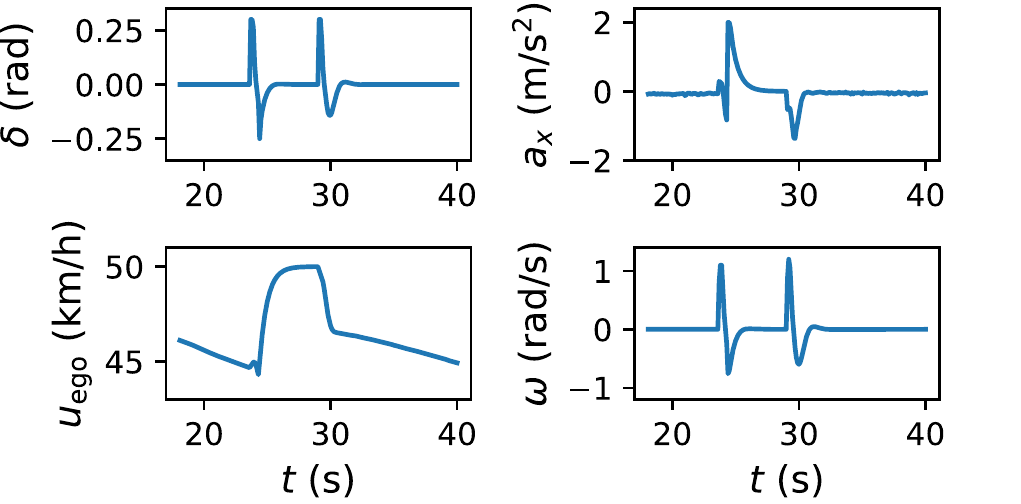}%
\label{fig:exp-var_init}}
%\hspace{-0.1in}
\subfloat[State and action variables of a typical\\lane-changing trajectory (After training)]{\includegraphics[width=0.35\linewidth, keepaspectratio=true,trim=45 560 60 40,clip]{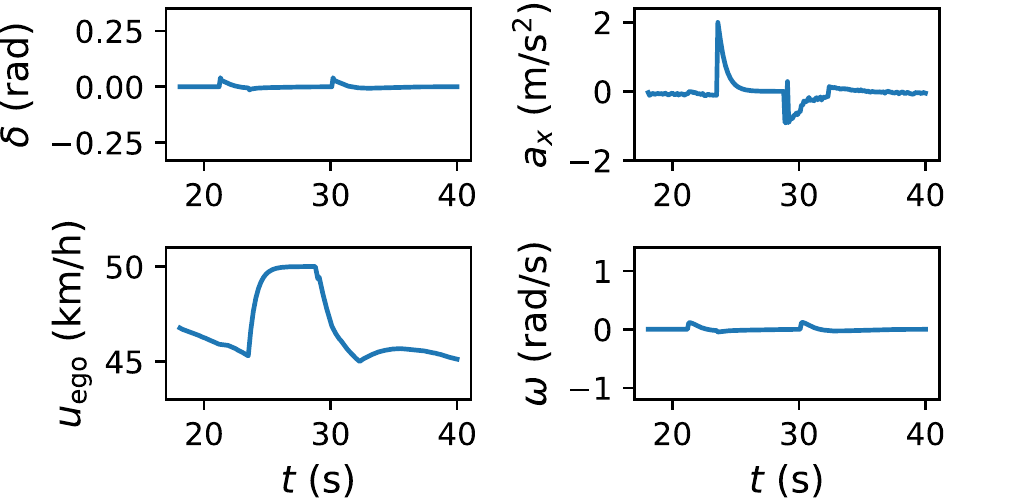}%
\label{fig:exp-var_learned}}
\vspace{0.01in}
\subfloat[Visualization of a typical lane-changing trajectory (Before training)]{\includegraphics[width=0.95\linewidth, keepaspectratio=true,trim=30 718 20 55,clip]{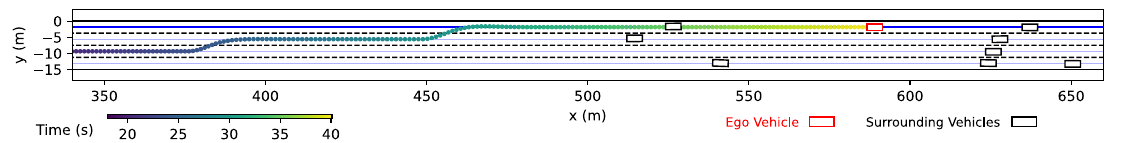}%
\label{fig:exp-traj_init}}
\vspace{0.01in}
\subfloat[Visualization of a typical lane-changing trajectory (After training)]{\includegraphics[width=0.95\linewidth, keepaspectratio=true,trim=30 718 20 55,clip]{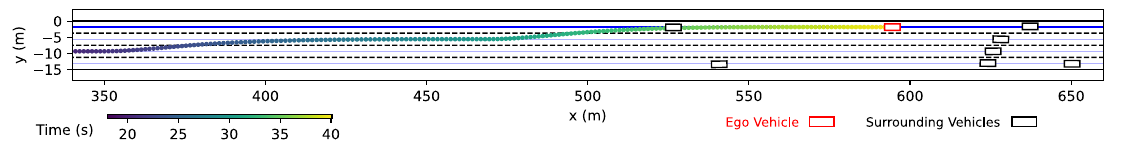}%
\label{fig:exp-traj_learned}}
%\hspace{-0.14in}
\caption{Visualization results of the driving policy before and after training with ZOAC.}
\vspace{-0.15in}
\label{fig:exp-vis}
\end{figure*}

Although all methods improve the performance over initial policy, ZOAC achieves the highest total average return of 5139.25. All autonomous driving policies drive at a similar average speed, but the one trained with ZOAC has the best riding comfort and fuel efficiency. Note that lane-changing process only constitutes a small proportion of the overall driving trajectory but contributes a lot to the riding comfort. One reason why ZOAC has gained an advantage possibly lies in the temporal credit assignment through actor-critic architecture, which distinguishes it from other methods that consider the entire trajectory as a whole.

Figure \ref{fig:exp-mean_evo} visualizes the evolution process of the quadratic weights used in the model predictive controller, where all weights are initialized to be the same and updated towards better driving performance. As presented in Figures \ref{fig:exp-var_init} and \ref{fig:exp-traj_init}, the ego vehicle performs two consecutive left lane-change and before driving along the innermost lane. However, the initial policy drives in a rough way, with large steering angle and yaw rate that lead to severe overshoot. This driving style is uncomfortable and does not meet practical needs. With performance-driven policy training, the cost weight on lateral position ($w_s^y$) decreases, while the cost weights on steering angle, lateral velocity, and yaw rate ($w_s^\delta$, $w_s^v$, $w_s^\omega$) increases. These changes significantly influence the driving behavior. From Figures \ref{fig:exp-var_learned} and \ref{fig:exp-traj_learned}, we can see that the ego vehicle conducts lane-changes with small steering angle and drives with improved safety and comfort.

\subsubsection{Ablation Study}
\label{sec:exp-ablation}

\begin{figure}[t]
\centering
% \subfloat[$N$ and $H$]{\includegraphics[width=0.25\textwidth, keepaspectratio=true,trim=32 360 68 46,clip]{figure/ablation_NH.pdf}%
% \label{fig:exp-ablation_NH}}
%\hspace{-0.1in}
\subfloat[Noise std. $\sigma$]{\includegraphics[height=1.45in, keepaspectratio=true,trim=35 380 120 50,clip]{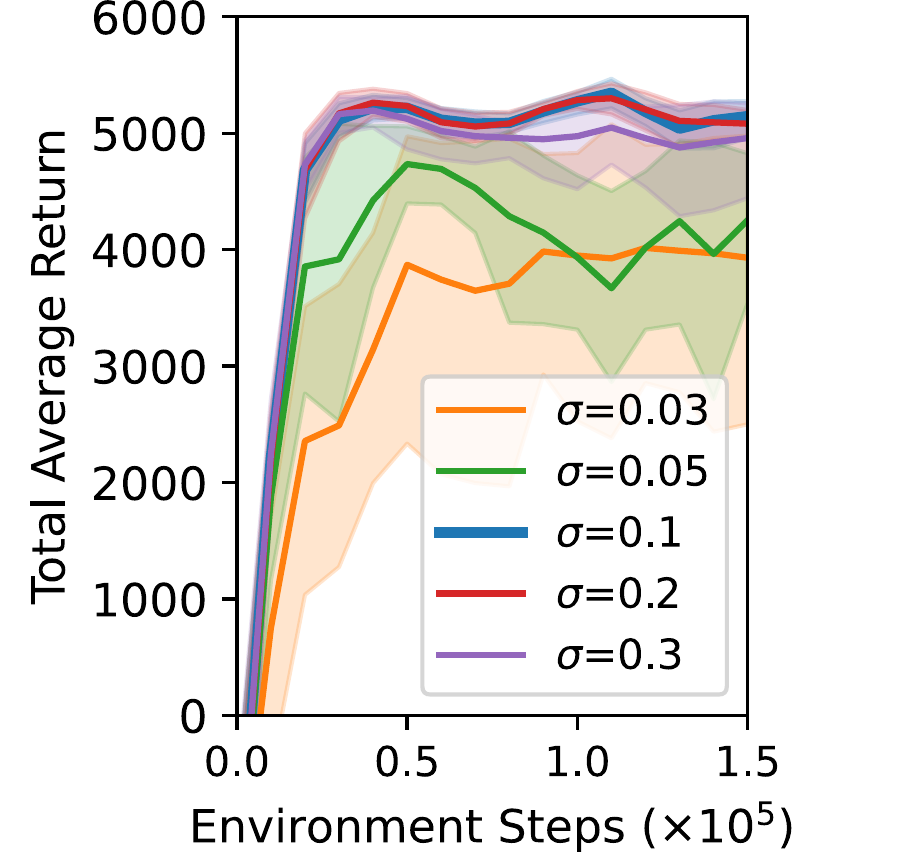}%
\label{fig:exp-ablation_sigma}}
\hspace{-0.24in}
\subfloat[GAE $\lambda$]{\includegraphics[height=1.42in, keepaspectratio=true,trim=71 380 165 48,clip]{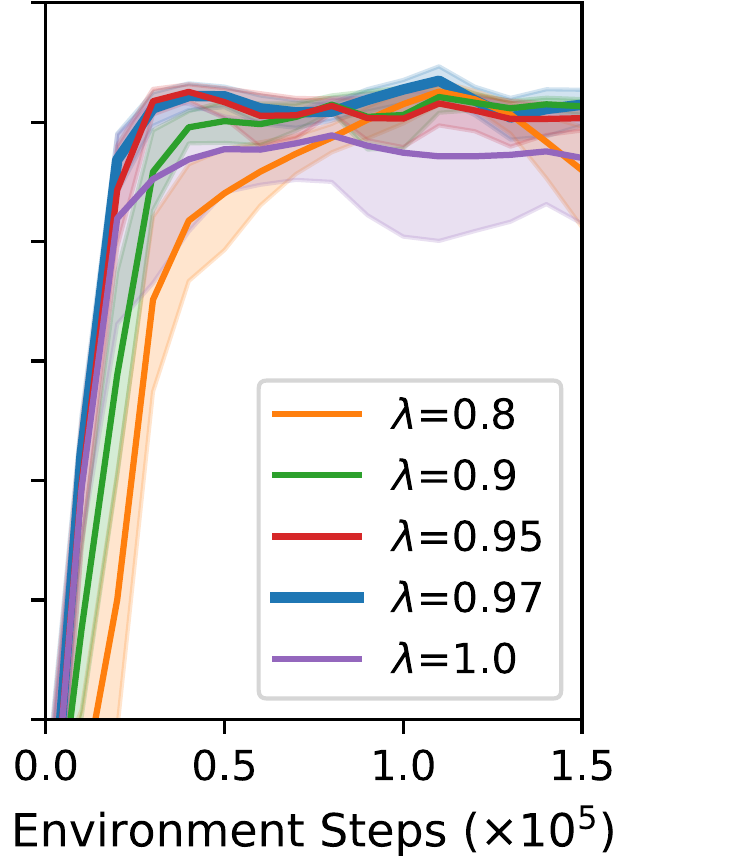}%
\label{fig:exp-ablation_gae}}
\hspace{-0.24in}
\subfloat[Learning rate $\alpha_{\text{actor}}$]{\includegraphics[height=1.42in, keepaspectratio=true,trim=71 380 215 48,clip]{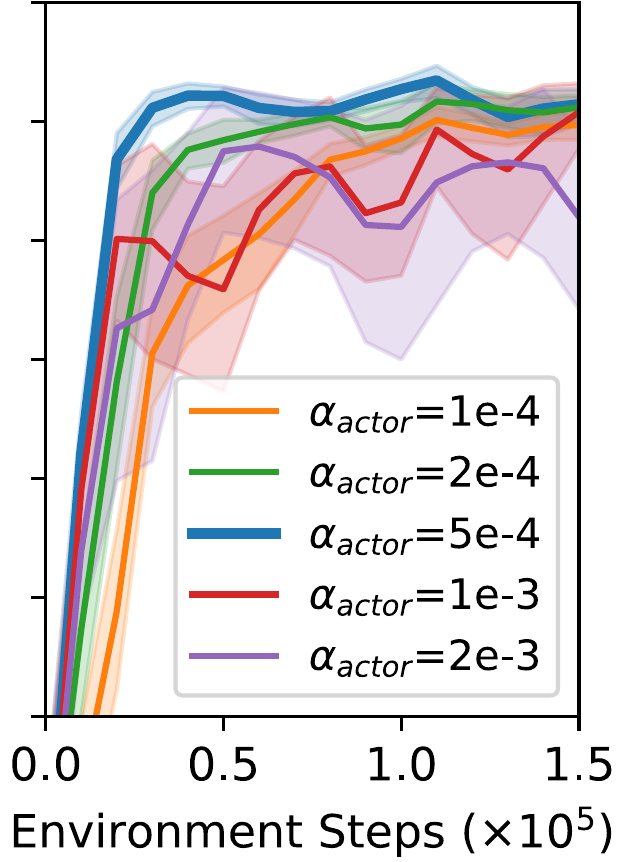}%
\label{fig:exp-ablation_lr}}
%\hspace{-0.1in}
\caption{Ablated learning curves of ZOAC in the autonomous driving task.}
\label{fig:exp-ablation_hyper}
\vspace{-0.15in}
\end{figure}

\begin{figure*}[t]
\centering
% \subfloat[$N$ and $H$]{\includegraphics[width=0.25\textwidth, keepaspectratio=true,trim=32 360 68 46,clip]{figure/ablation_NH.pdf}%
% \label{fig:exp-ablation_NH}}
%\hspace{-0.1in}
\subfloat[Illustration of Gymnasium tasks]{\includegraphics[height=1.9in, keepaspectratio=true,trim=170 130 170 130,clip]{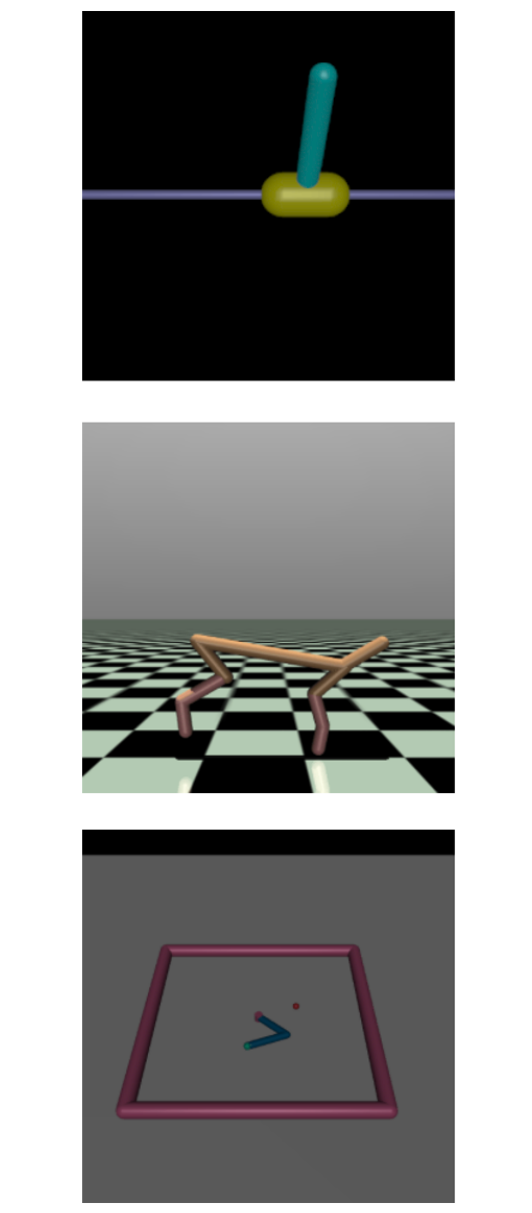}%
\label{fig:exp-mujoco-env}}
\hspace{0.15in}
\subfloat[InvertedPendulum-v4\\($|\mathcal{S}|=4$, $|\mathcal{A}|=1$, $D=4545$)]{\includegraphics[height=1.9in, keepaspectratio=true,trim=50 280 48 46,clip]{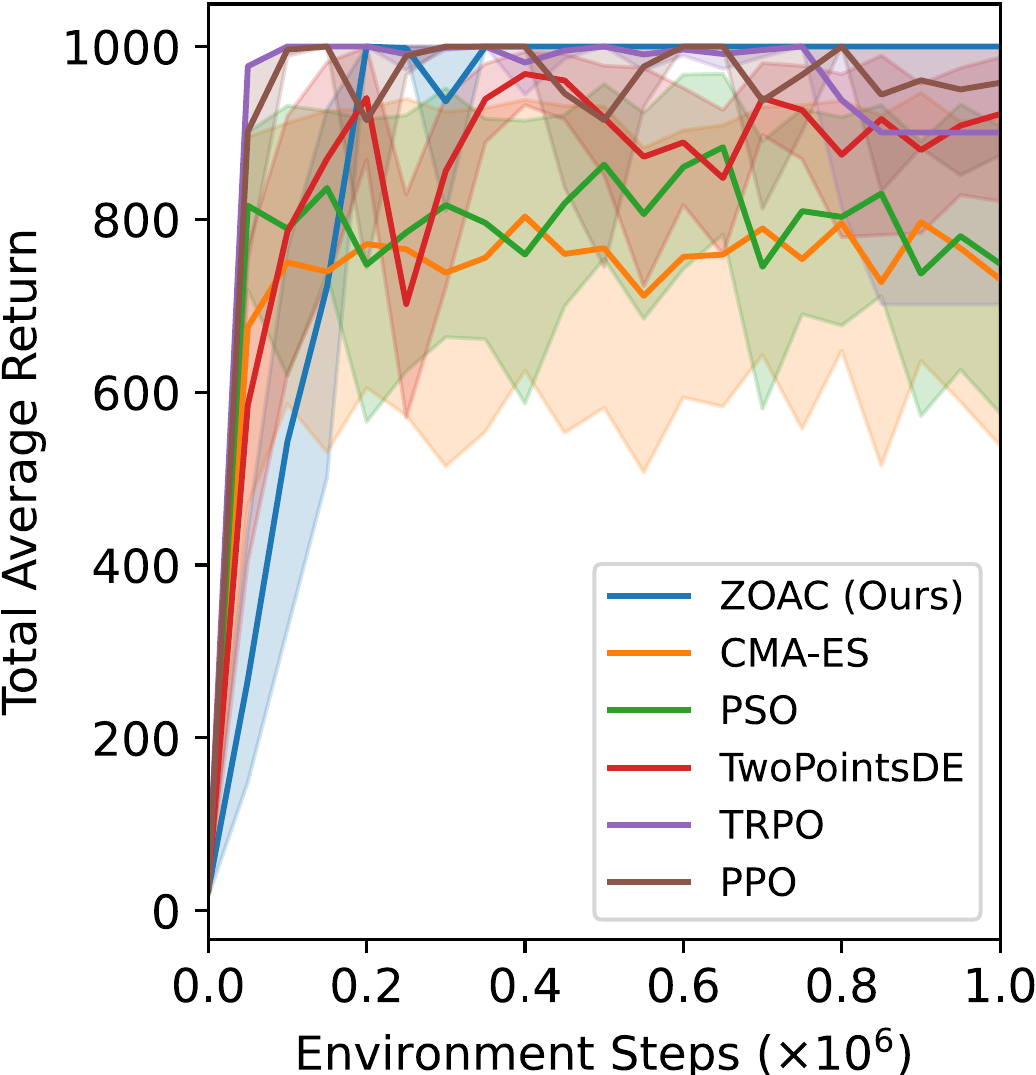}%
\label{fig:exp-mujoco-invp}}
\hspace{0.1in}
\subfloat[HalfCheetah-v4\\($|\mathcal{S}|=17$, $|\mathcal{A}|=6$, $D=5702$)]{\includegraphics[height=1.9in, keepaspectratio=true,trim=50 280 48 46,clip]{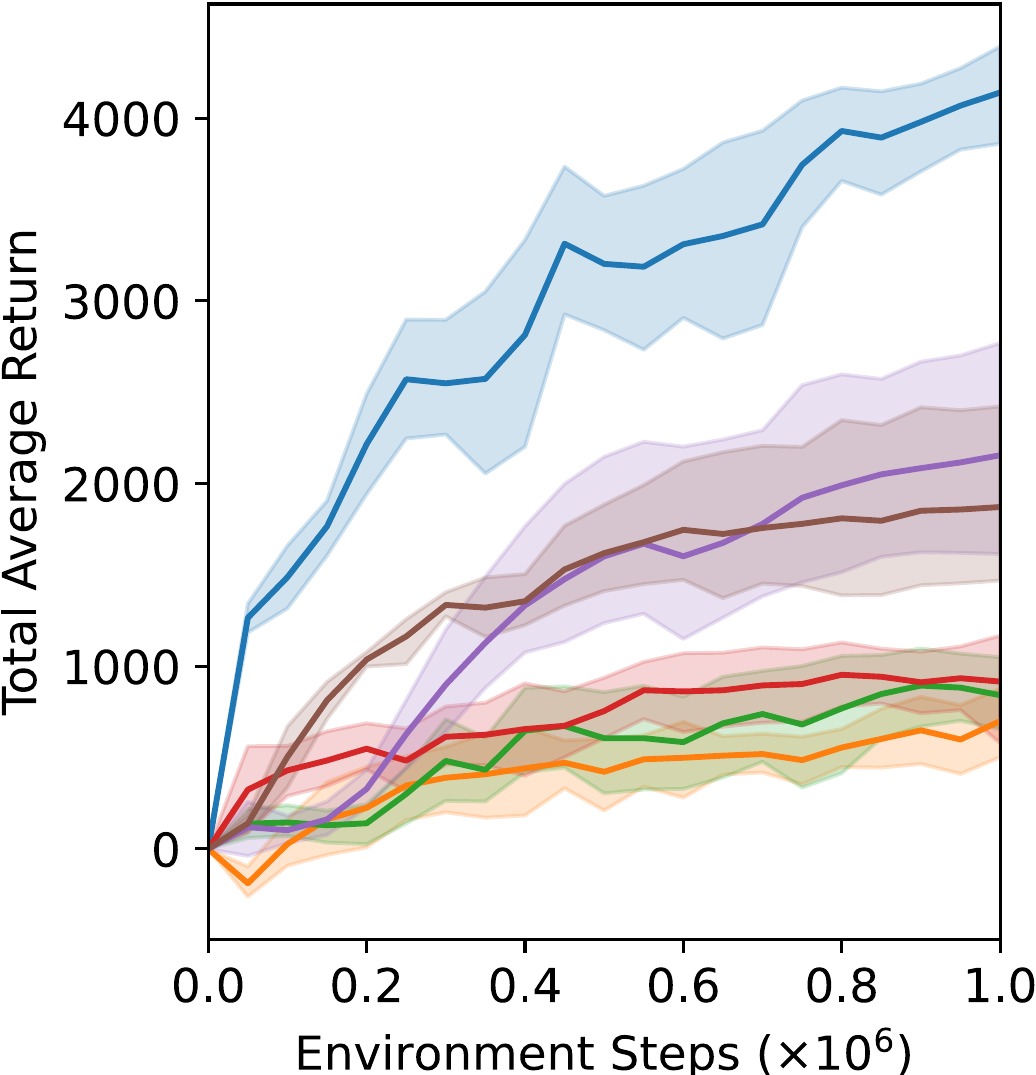}%
\label{fig:exp-mujoco-hfc}}
\hspace{0.1in}
\subfloat[Reacher-v4\\($|\mathcal{S}|=11$, $|\mathcal{A}|=2$, $D=5058$)]{\includegraphics[height=1.9in, keepaspectratio=true,trim=50 280 48 46,clip]{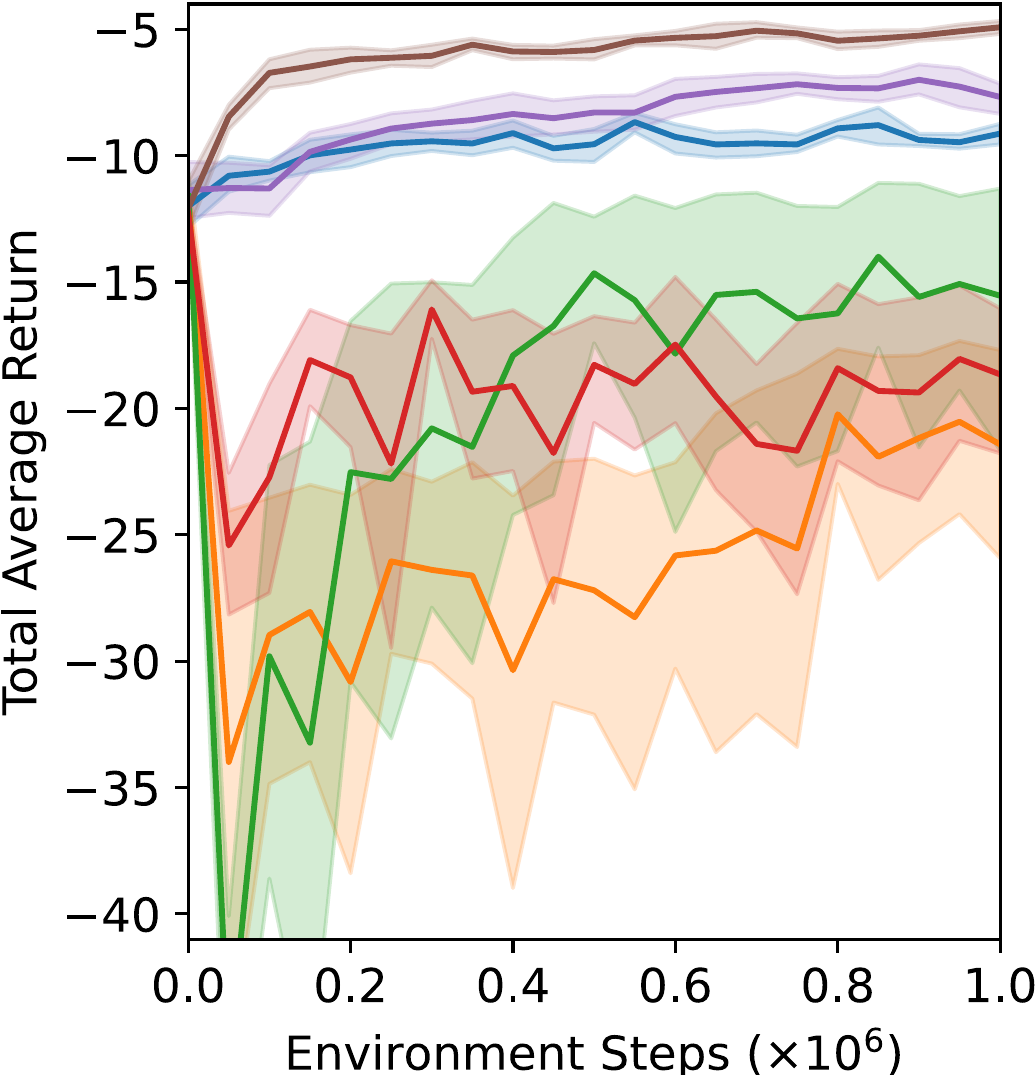}%
\label{fig:exp-mujoco-rc}}
%\hspace{-0.1in}
\caption{Learning curves of ZOAC, zeroth-order optimization and gradient-based RL baselines in three Gymnasium tasks: InvertedPendulum-v4, HalfCheetah-v4, Reacher-v4. (a) illustrates the environments from top to down respectively, while (b)-(d) shows the learning curves over 10 random seeds. We also provide the dimensions of the state space $\mathcal{S}$, the action space $\mathcal{A}$, and the neural network policies.}
\vspace{-0.15in}
\label{fig:exp-mujoco}
\end{figure*}

In Section \ref{sec:framework.zopg}, Theorem \ref{theorem.bound} analyze the variance of ES and ZOAC gradient estimators. As the value of $N$ decreases, the \textit{fitness} values become more dependent on the critic network, resulting in smaller variance but potentially larger bias, and vice versa. This suggests that using an appropriate rollout length $N$ of each perturbed policies can help strike a balance between bias and variance. We conduct an ablation study to understand the influence of $N$ under the condition that the budget of timesteps within one iteration is 1,000 ((i.e., $N \times H$ equals to the episodic length). The autonomous driving policy is updated by either \eqref{eq:prel-esgradest.mean} or \eqref{eq:method-zoacadvest.mean}. As shown in Figure \ref{fig:exp-ablation-NH}, incorporating actor-critic architecture leads to significant improvement on efficiency and stability compared to ES, among which ZOAC with $N=40,H=25$ performs the best. We also ablate ZOAC on other important hyperparameters in Figure \ref{fig:exp-ablation_hyper}, including Gaussian noise standard deviation $\sigma$, GAE coefficient $\lambda$, and learning rate $\alpha_{\text{actor}}$. We modify each hyperparameter individually while keeping the others unchanged as in Table \ref{tab:hyper}. Our findings suggest that a too small $\sigma$ may lead to suboptimal solutions, while a moderate learning rate can ensure both high learning speed and stability. Nonetheless, ZOAC is generally not sensitive to the choice of hyperparameters, making it easy to use in practice.

Since ZOAC shows significant improvement over ES by incorporating temporal difference learning, we wonder if the hybridization can be applied to other baselines that also use Monte Carlo returns like ES. We apply a similar actor-critic architecture to other baselines, where each candidate rolls out forward $N=40$ steps and its fitness is estimated as the state-value target $\hat{G}$ in \eqref{eq:method-prac.tarest}. However, Figure \ref{fig:exp-ablation-hybridbaseline} show that it performs poorly. We hypothesize that this heuristic hybridization fails because, in ES, the population at iteration $k+1$ only depends on the fitness of candidates that are close to each other from the $k$-th iteration. In contrast, other evolutionary algorithms might perform mutation or crossover over the entire search space in one iteration and incorporate candidates from older iterations (e.g., historically best candidates). Since the critic network only accurately approximates the state-value function of the current policy, the outdated state values stored in the algorithm can be seriously misleading. A tailored hybridization approach might be needed for each specific method, which we leave for future exploration.

We also examine the computational cost of hybridization in Figure \ref{fig:exp-ablation-time}. With hyperparameters $N=40$ and $H=25$, all hybrid variants require approximately 150 additional seconds to reach 3e5 steps, primarily due to the training of the additional critic network. Additionally, the training time increases with the population size because of the higher cost associated with resampling policies and transferring transition data between the learner and samplers. Although the overall training time for ZOAC is longer, its convergence time (here we refer to the time to achieve a total average return exceeding 4,000) is much faster than the zeroth-order baselines.

\subsection{Additional Results in Gymnasium Tasks}
\label{sec:exp-gym}

Although ZOAC is primarily designed for optimizing non-differentiable policies, we are still curious about its performance on the standard Gymnasium benchmark \cite{towers2024gymnasium}, especially in comparison with gradient-based RL algorithms. To this end, we use two popular actor-critic baselines from the open-source library GOPS \cite{wang2023gops}: (1) Trust Region Policy Optimization (TRPO) \cite{schulman2015trust}, which enforces a KL divergence constraint on the policy update size, and (2) Proximal Policy Optimization (PPO) \cite{schulman2017proximal}, which simplifies the constraint in TRPO by using a clipped surrogate objective. the policies to be learned are chosen as standard two-layer MLPs with 64 hidden nodes and tanh nonlinearities. The critic network remains the same as in the previous section for all three actor-critic methods. According to the default settings in GOPS, both TRPO and PPO use $n=8$ parallel samplers, each collecting 128 transition data samples in an iteration (i.e., 1024 samples per iteration). To align with this, we set $N=8$ and $H=16$ in ZOAC experiments, ensuring that each sampler collects the same number of 128 samples per iteration. Besides, we use smaller Gaussian noise $\sigma=0.05$ for neural network policies. All other hyperparameters remain consistent with those used in the autonomous driving task. 

The learning curves in Figure \ref{fig:exp-mujoco} indicate that, although ZOAC does not require backpropagating gradients in the policies, it generally matches the performance of gradient-based RL methods and significantly outperforms zeroth-order baselines across three benchmark tasks. This result highlights the importance of incorporating temporal difference learning and the actor-critic architecture, enabling ZOAC to handle sequential decision problems with efficiency comparable to gradient-based RL methods like TRPO and PPO, regardless of whether the policies are differentiable.

\section{Conclusion}
\label{sec:conclusion}

This paper proposes an evolutionary framework called the Zeroth-Order Actor-Critic algorithm (ZOAC) to address sequential decision problems. ZOAC combines policy gradient and actor-critic architecture with evolutionary computation, effectively leveraging the inherent temporal structure to enhance sample efficiency and training stability.
In each iteration, ZOAC employs parallelized samplers to collect rollouts with timestep-wise perturbation in parameter space. The learner then alternates between first-order policy evaluation (PEV) and zeroth-order policy improvement (PIM). Experimental results on a challenging multi-lane autonomous driving task and three Gymnasium tasks demonstrate the superior performance of ZOAC over existing evolutionary algorithm baselines that treat the problem as static optimization.

Future research can explore several directions. Firstly, extending the framework to optimize parameters in a discrete space would expand its applicability to a wider range of settings. Besides, integrating exploration and updating schemes that consider safety constraints into the ZOAC framework would be beneficial for practical applications. Finally, while the proposed algorithms have been extensively validated in simulation, conducting further validation on dynamic systems in real environments could be more convincing.

\bibliographystyle{IEEEtran}
\bibliography{paper}

% \newpage

%\vspace{-0.5in}
\begin{IEEEbiography}
[{\includegraphics[width=1in,height=1.25in,clip,keepaspectratio]{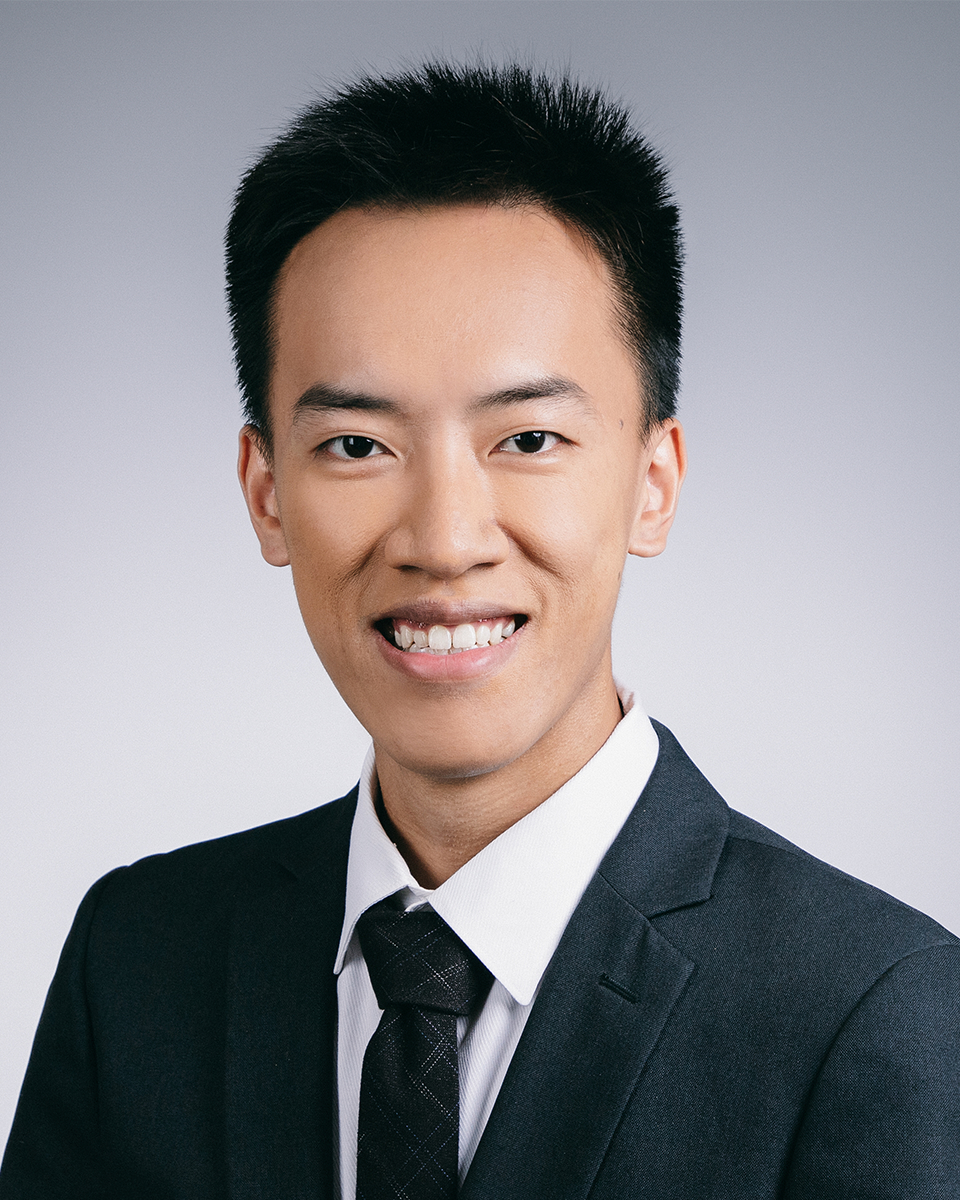}}]{Yuheng Lei} received the B.E. and M.Sc. degrees from Tsinghua University, in 2020 and 2023, respectively. He is currently pursuing the Ph.D. degree in Computer Science at the University of Hong Kong. His current research interests include embodied AI, reinforcement learning, robotic control, and autonomous driving.
\end{IEEEbiography}
\vspace{-0.5in}
\begin{IEEEbiography}
[{\includegraphics[width=1in,height=1.25in,clip,keepaspectratio]{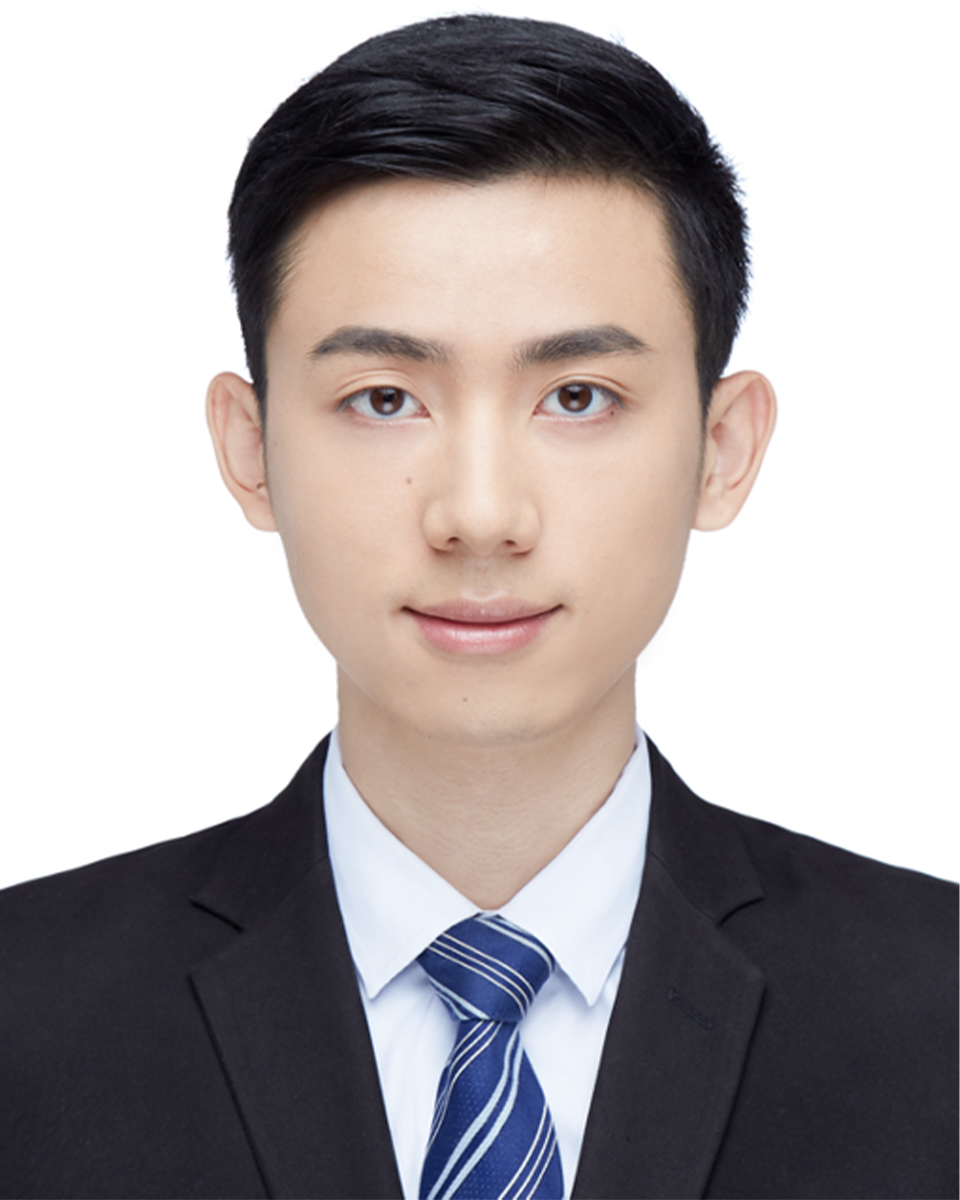}}]{Yao Lyu} received his B.E. degree in Automotive Engineering from Tsinghua University, Beijing, China, in 2019. He is currently pursuing his Ph.D. degree in Automotive Engineering from School of Vehicle and Mobility, Tsinghua University, Beijing, China. His active research interests include decision-making and control of autonomous vehicles, reinforcement learning, accelerating optimization and optimal control theory.
\end{IEEEbiography}

\begin{IEEEbiography}
[{\includegraphics[width=1in,height=1.25in,clip,keepaspectratio]{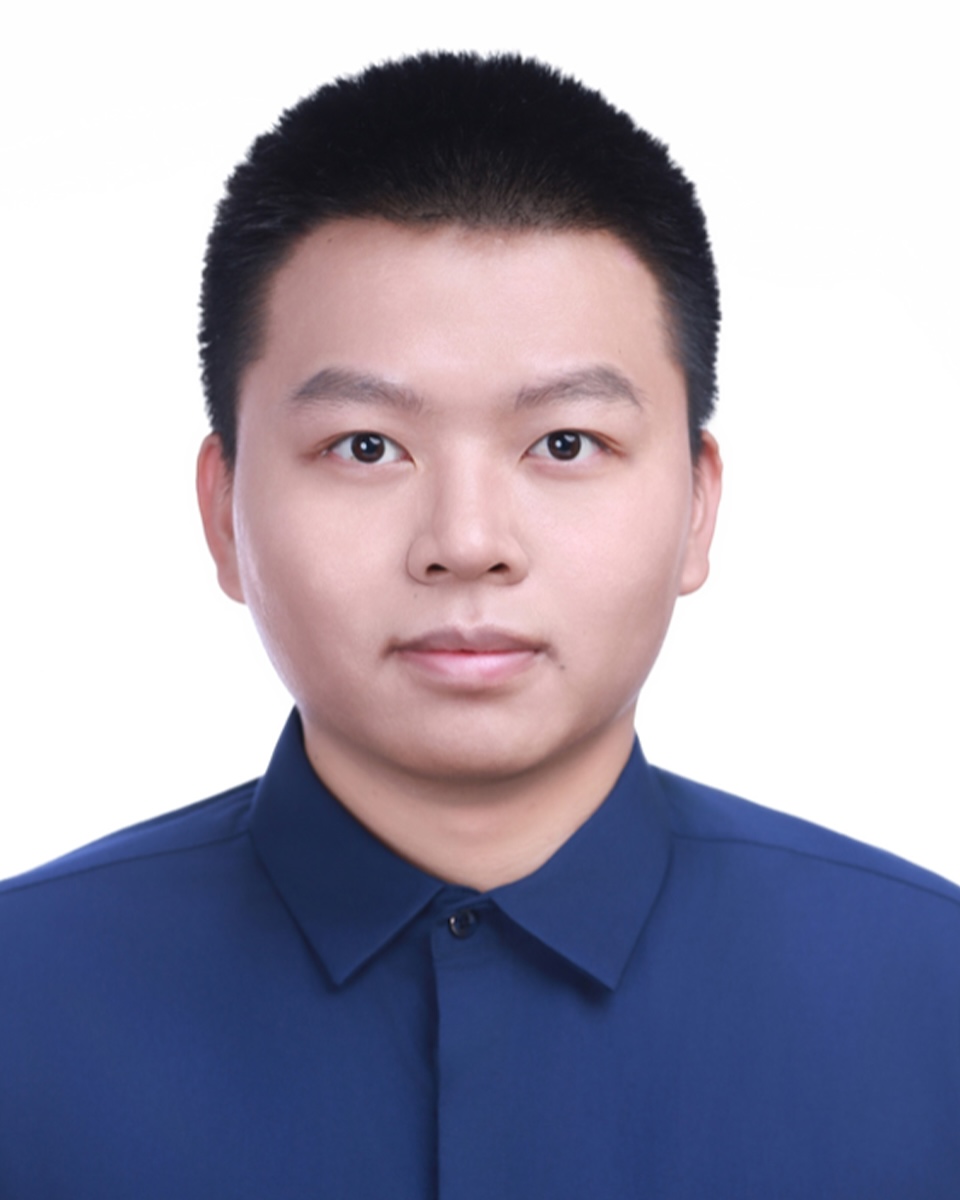}}]{Guojian Zhan} received the B.E. degree from the School of Vehicle and Mobility, Tsinghua University, Beijing, China, in 2021. He is pursuing a Ph.D. in Mechanical Engineering at the School of Vehicle and Mobility, Tsinghua University. His research focuses on control theory, reinforcement learning, and their applications in autonomous vehicles and robotics.
\end{IEEEbiography}
\vspace{-0.5in}
\begin{IEEEbiography}
[{\includegraphics[width=1in,height=1.25in,clip,keepaspectratio]{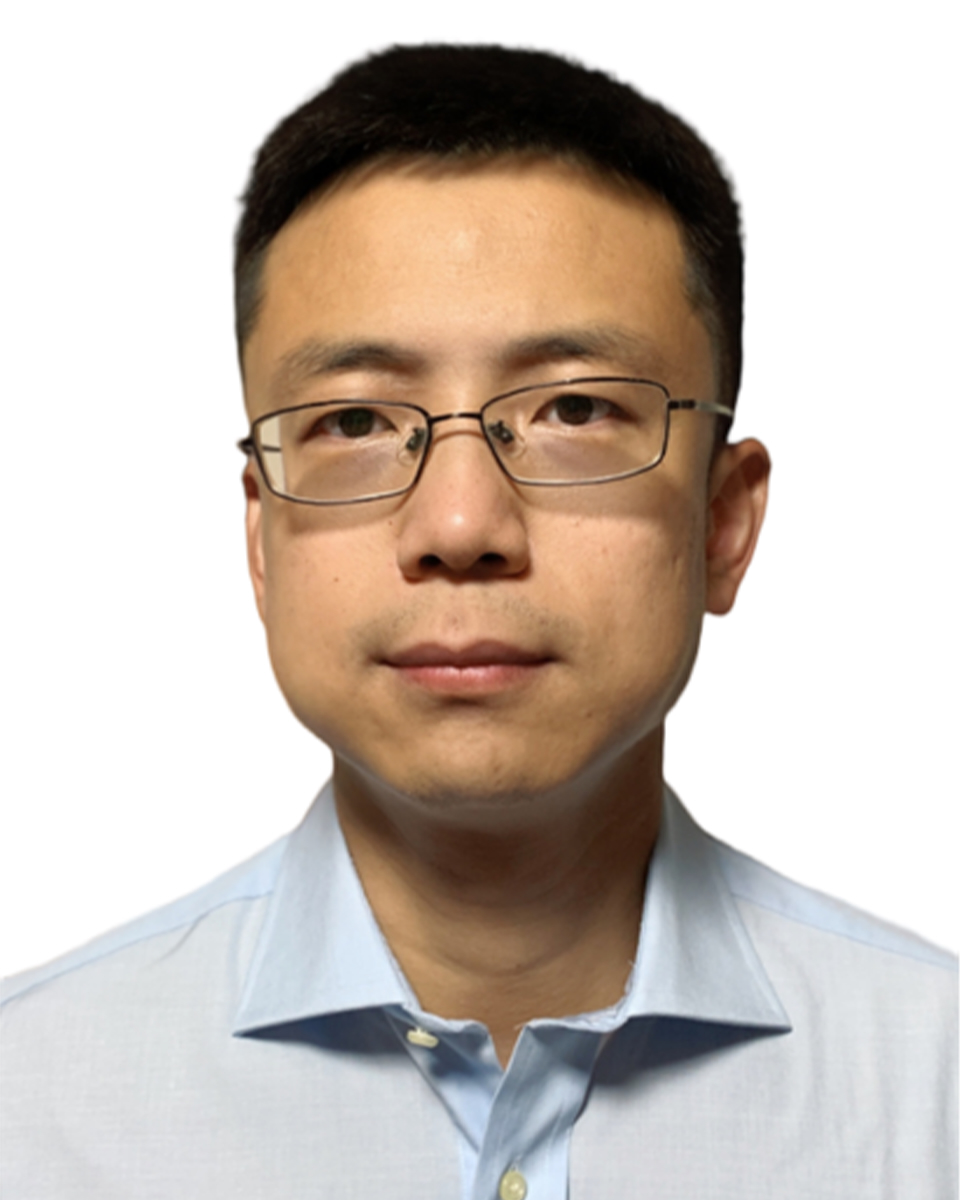}}]{Tao Zhang} received his B.S. and Ph.D. degrees from Tsinghua University in 2005 and 2010. Before joining Autonavi, Alibaba Group, he has worked at Politecnico di Milano as an independent researcher. His active research interests include intelligent vehicles, localization and navigation systems, optimal estimation, and SLAM. He led the sensor and positioning group at Autonavi, which won first place in the IPIN positioning competition for three consecutive years (2020, 2021, 2022), and second place in the 2024 CVPR Image Matching Challenge. He also led the development of the world’s first lane-level navigation system and a popular AR application with millions of active users.
\end{IEEEbiography}
\vspace{-0.5in}
\begin{IEEEbiography}
[{\includegraphics[width=1in,height=1.25in,clip,keepaspectratio]{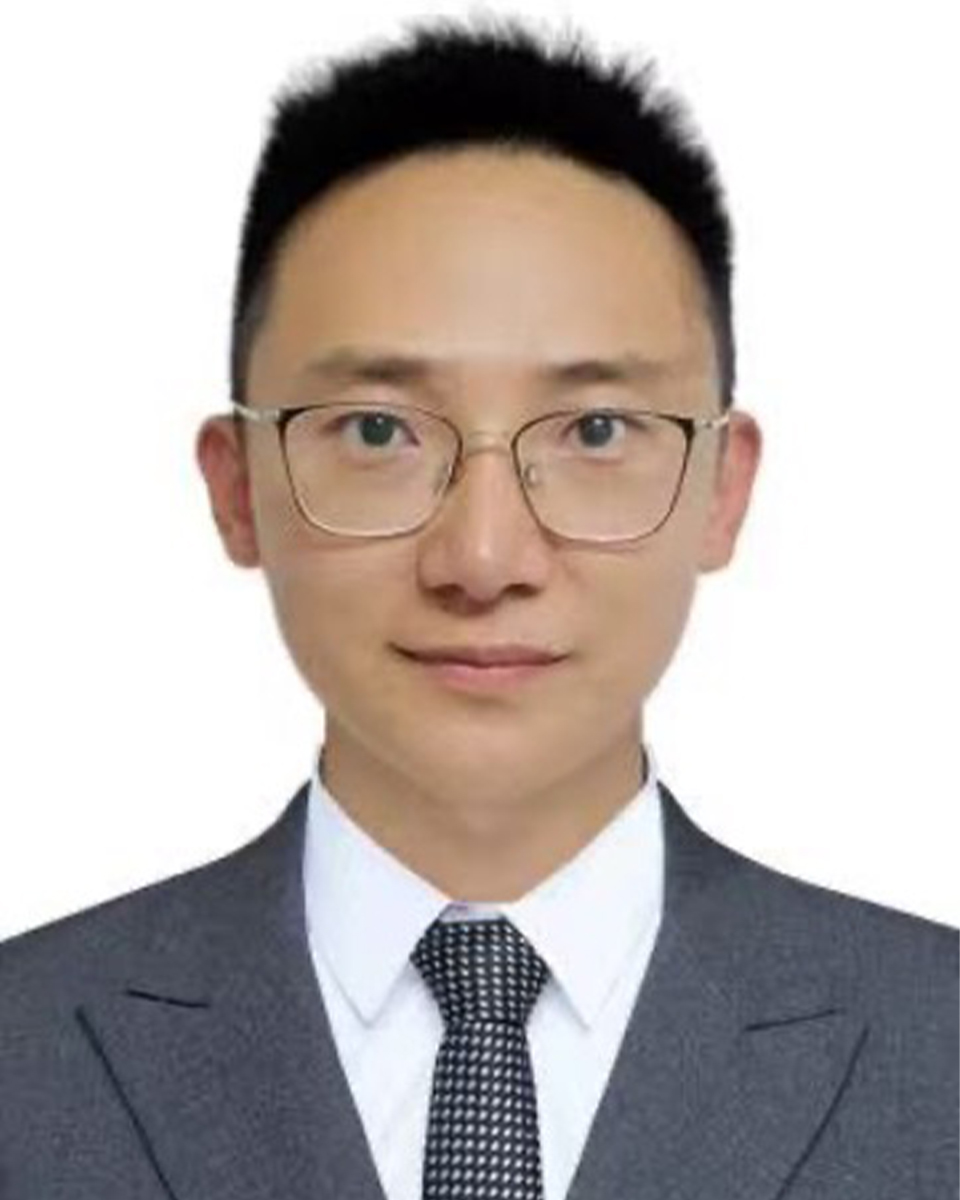}}]{Jiangtao Li} received his Ph.D. degree from Tsinghua University in 2015 and joined Autonavi, Alibaba Group in 2016. His active research interests include lane-level localization and navigation systems, machine learning, etc. He led the lane-level positioning group in Autonavi and developed the world’s first lane-level navigation system for both mobile phones and vehicles.
\end{IEEEbiography}
\vspace{-0.5in}
\begin{IEEEbiography}
[{\includegraphics[width=1in,height=1.25in,clip,keepaspectratio]{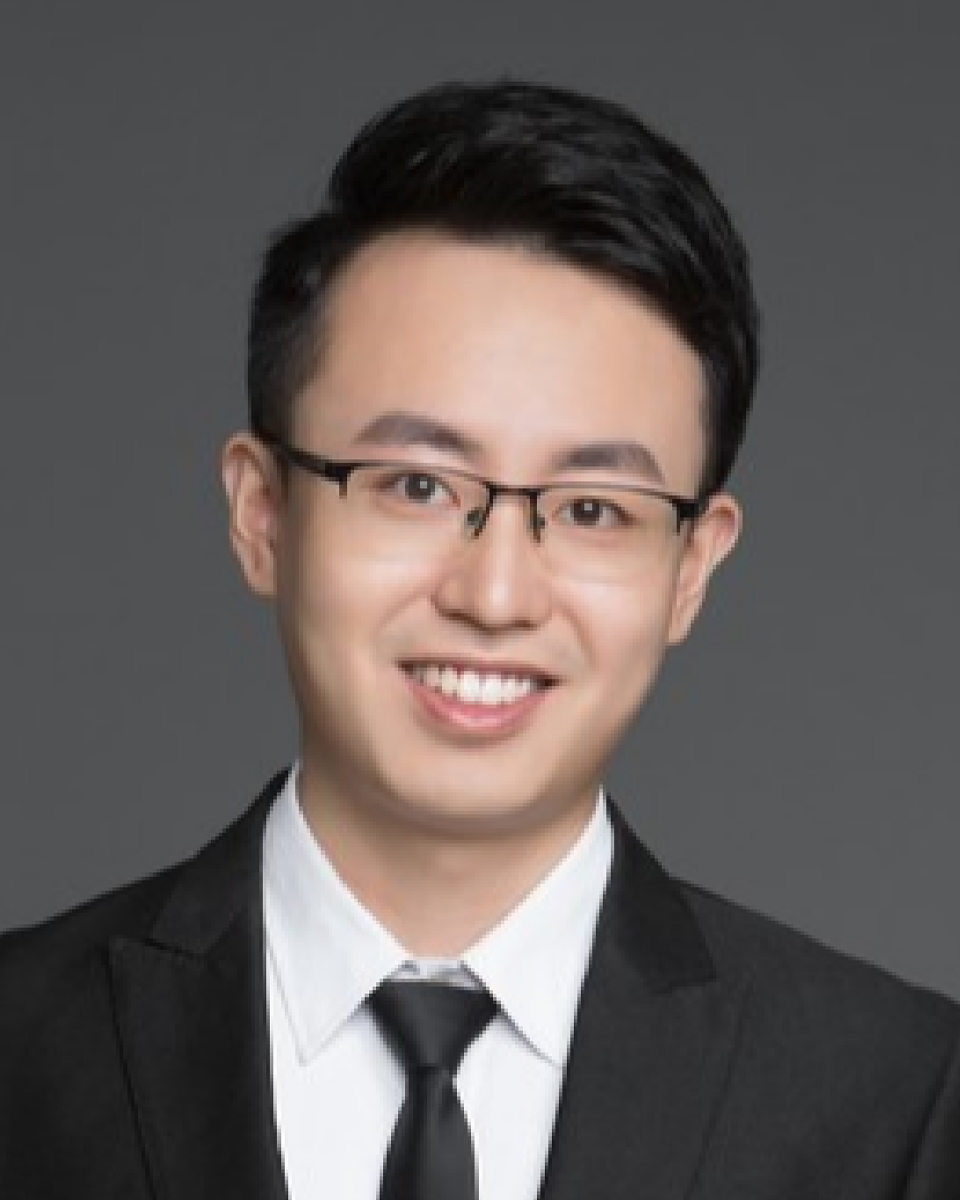}}]{Jianyu Chen} received the B.E. degree from Tsinghua University, China, in 2015. He received the Ph.D. degree working with Prof. Masayoshi Tomizuka at the University of California, Berkeley in 2020. He is currently an Assistant Professor with the Institute for Information Sciences (IIIS), Tsinghua University. He is working in the cross fields of robotics, reinforcement learning, control, and autonomous driving. His research goal is to build advanced robotic systems with high performance and high intelligence.
\end{IEEEbiography}
\vspace{-0.5in}
\begin{IEEEbiography}
[{\includegraphics[width=1in,clip,keepaspectratio]{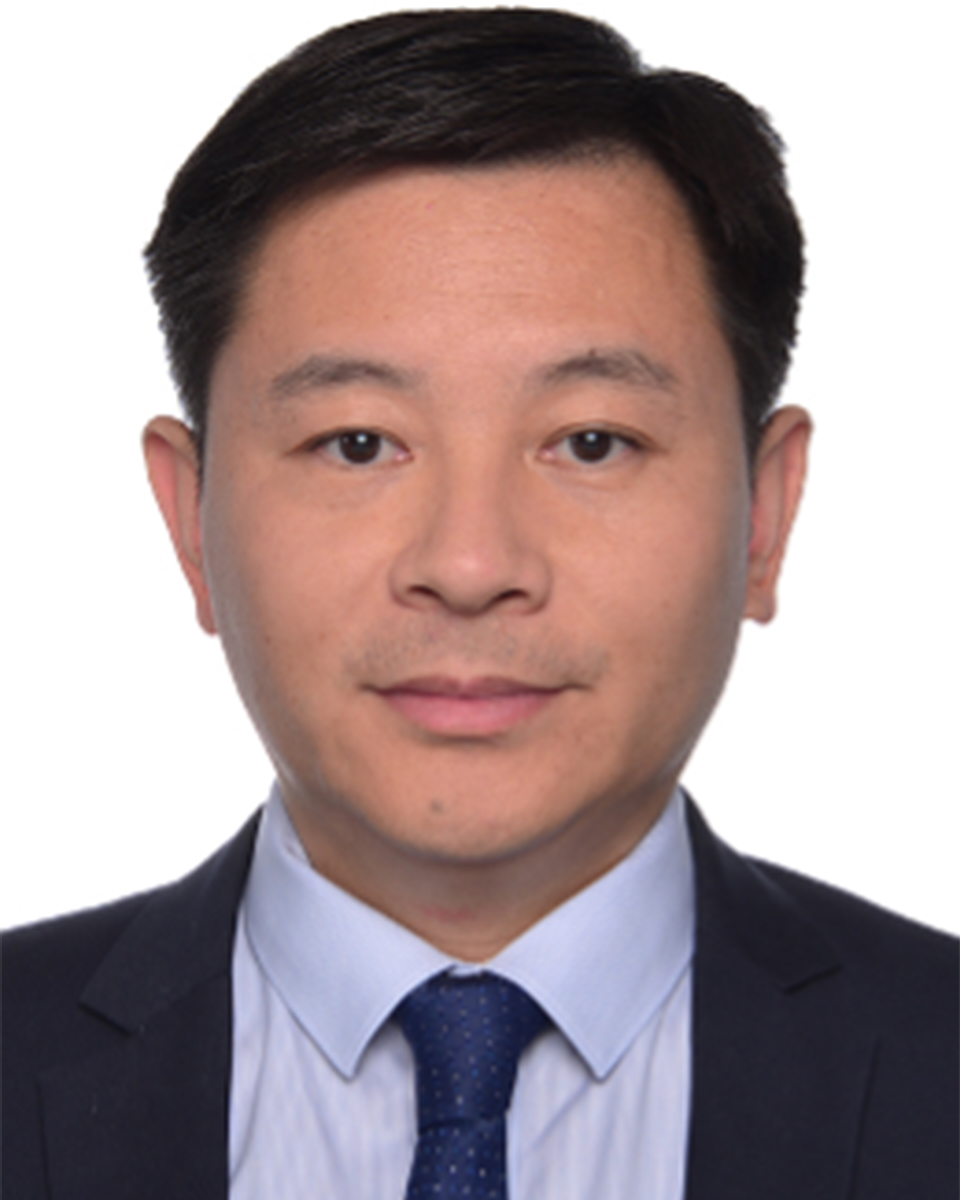}}]{Shengbo Eben Li} (Senior Member, IEEE) received his M.S. and Ph.D. degrees from Tsinghua University in 2006 and 2009. Before joining Tsinghua University, he has worked at Stanford University, University of Michigan, and UC Berkeley. His active research interests include intelligent vehicles and driver assistance, deep reinforcement learning, optimal control and estimation, etc. He is the author of over 190 peer-reviewed journal/conference papers, and co-inventor of over 40 patents. Dr. Li has received over 20 prestigious awards, including Youth Sci. \& Tech Award of Ministry of Education (annually 10 receivers in China), Natural Science Award of Chinese Association of Automation (First level), National Award for Progress in Sci \& Tech of China, and best (student) paper awards of IET ITS, IEEE ITS, IEEE ICUS, CVCI, etc. He also serves as Board of Governor of IEEE ITS Society, Senior AE of IEEE OJ ITS, and AEs of IEEE ITSM, IEEE TITS, IEEE TIV, IEEE TNNLS, etc. 
\end{IEEEbiography}
\vspace{-0.5in}
% insert where needed to balance the two columns on the last page with
% biographies
%\newpage
\begin{IEEEbiography}
[{\includegraphics[width=1in,height=1.25in,clip,keepaspectratio]{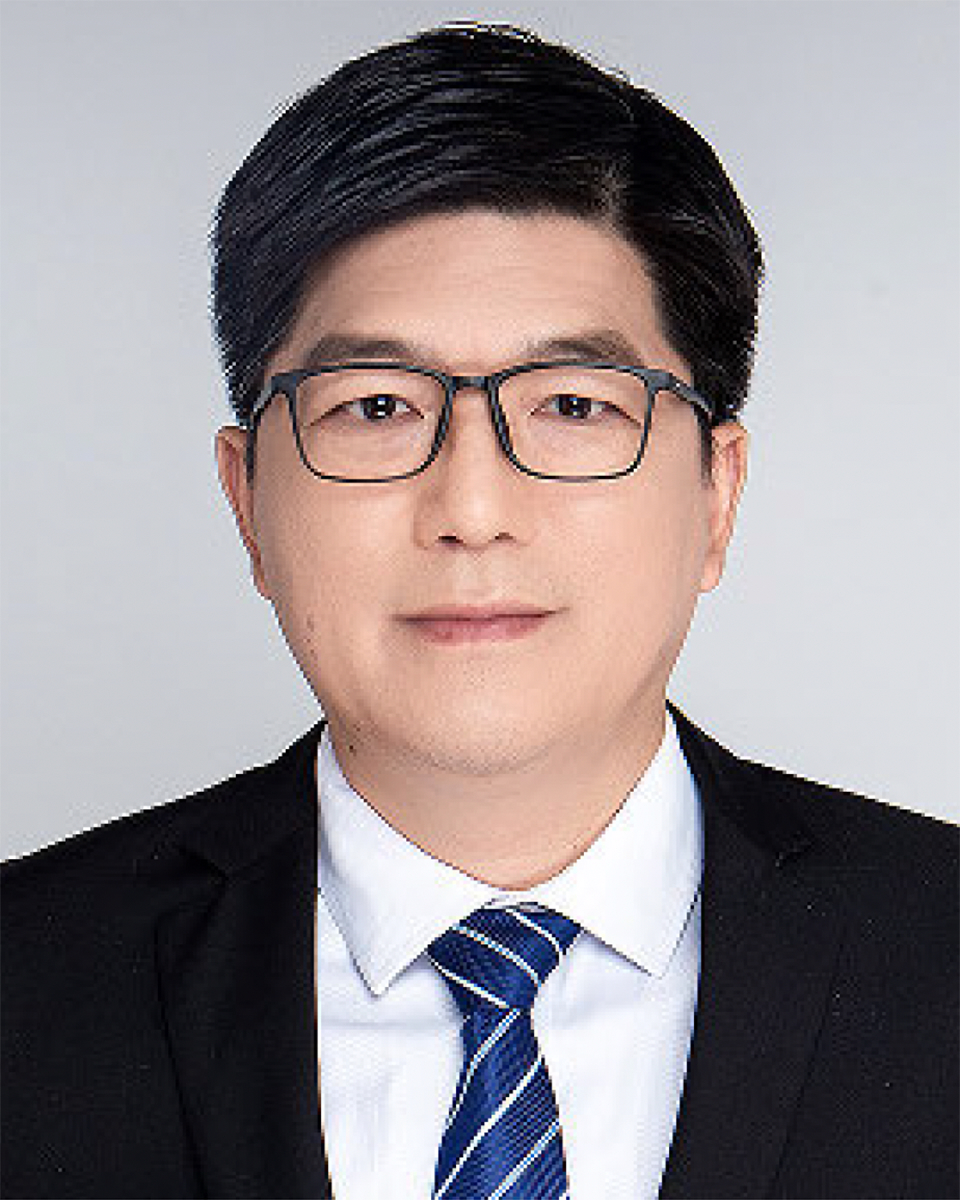}}]{Sifa Zheng} (Member, IEEE) received the B.E. and Ph.D. degrees from Tsinghua University, Beijing, China, in 1993 and 1997, respectively. He is currently a Professor with the State Key Laboratory of Automotive Safety and Energy, School of Vehicle and Mobility, Tsinghua University, where he is also the Deputy Director of the Suzhou Automotive Research Institute. His current research interests include autonomous driving and vehicle dynamics and control. 
\end{IEEEbiography}

\includepdf[pages=-]{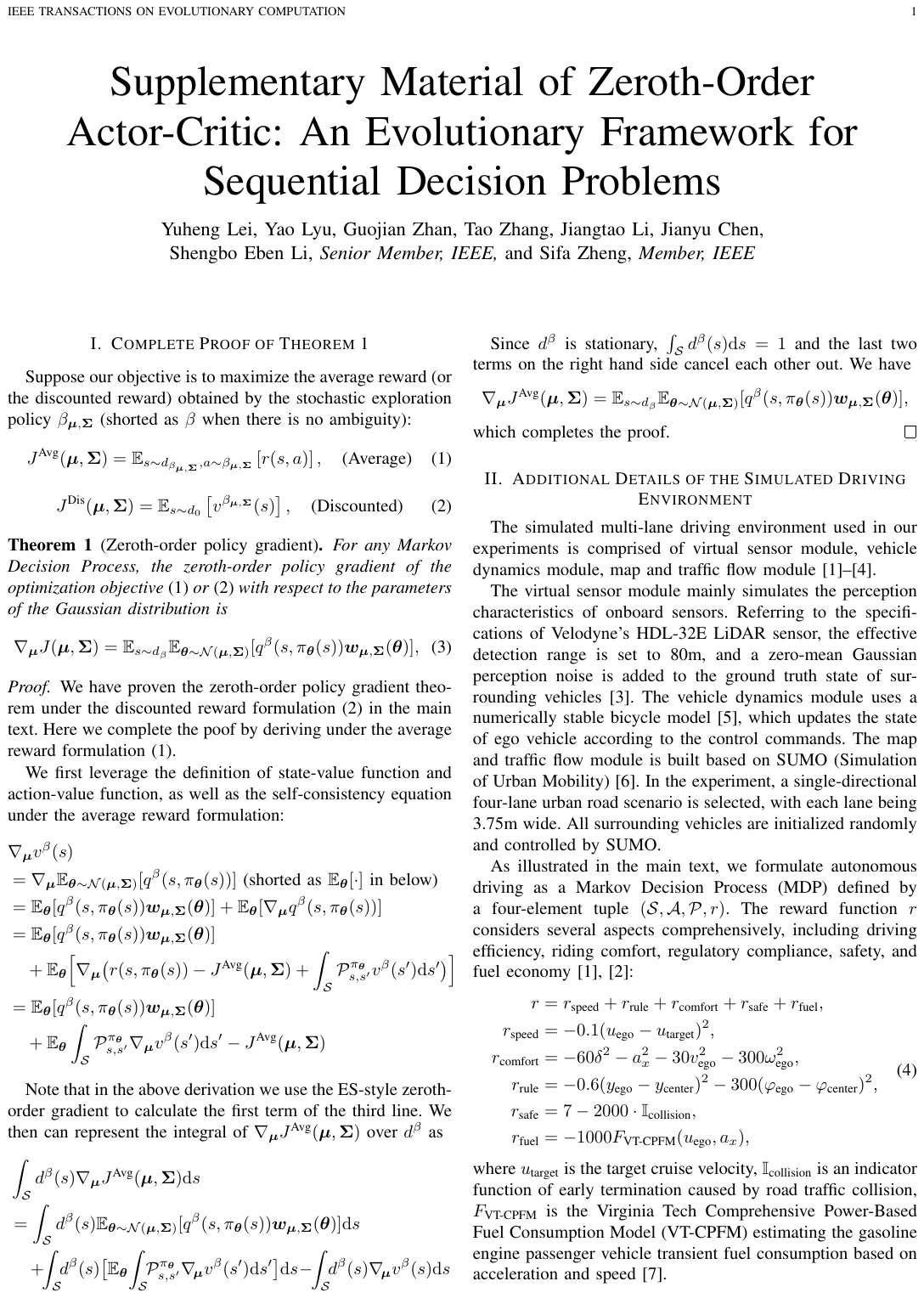}

\end{document}